\documentclass[final,5p,times,twocolumn]{elsarticle}

\usepackage{amsmath,amsfonts,amssymb}
\usepackage{bm}
\usepackage{mathtools}
\usepackage{subcaption}
\usepackage{algorithm}
\usepackage{algpseudocode}
\usepackage{array}
\usepackage{booktabs}
\usepackage{tabularx}
\usepackage{multirow}
\usepackage{graphicx}
\graphicspath{{./}{figures/}}
\usepackage{stfloats}
\usepackage[utf8]{inputenc}
\usepackage{textcomp}
\usepackage{url}
\usepackage{verbatim}
\usepackage{pifont}
\usepackage[dvipsnames]{xcolor}
\newcommand{\cmark}{\textcolor{ForestGreen}{\ding{51}}}
\newcommand{\xmark}{\textcolor{BrickRed}{\ding{55}}}
\usepackage{bbm}
\usepackage{enumitem}
\usepackage{balance}
\usepackage{comment}
\usepackage{tikz}
\usetikzlibrary{shapes,arrows,positioning,trees,mindmap}
\usepackage[switch]{lineno}
\usepackage{placeins}
\usepackage{float}
\usepackage[colorlinks=true,
            linkcolor=blue,
            citecolor=blue,
            urlcolor=blue]{hyperref}

\newcommand{\best}[1]{\textbf{#1}}

\newcommand{\nd}{\textemdash}

\biboptions{numbers,sort&compress}
\journal{}

\begin{document}
\begin{frontmatter}

\title{Does Explainability Transfer? A Controlled Benchmark of Attribution Methods on Vision Transformers and CNNs}

\author[aff1]{Sathiyamohan Nishankar\corref{cor1}}
\ead{e17230@eng.pdn.ac.lk}
\author[aff2]{Nethmi Pathirana}
\author[aff3]{Pubudu Sanjeewani}
\author[aff4,aff5]{Asanka Perera}
\author[aff6]{Selvarajah Thuseethan}

\cortext[cor1]{Corresponding author}

\affiliation[aff1]{organization={Faculty of Computing, Sabaragamuwa University of Sri Lanka},
            country={Sri Lanka}}
\affiliation[aff2]{organization={Faculty of Engineering, University of Moratuwa},
            city={Katubedda},
            country={Sri Lanka}}
\affiliation[aff3]{organization={School of Computing Technologies, RMIT University},
            city={Melbourne},
            country={Australia}}
\affiliation[aff4]{organization={School of Engineering \& Digital Technologies, University of Southern Queensland},
            city={Brisbane},
            country={Australia}}
\affiliation[aff5]{organization={School of Engineering \& Technologies, UNSW},
            city={Canberra},
            country={Australia}}            
\affiliation[aff6]{organization={Faculty of Science and Technology, Charles Darwin University, Australia},
            city={Darwin},
            country={Australia}}

\begin{abstract}
Most evidence on the effectiveness of explainable artificial intelligence (XAI) attribution methods has been established on convolutional neural networks (CNNs), with limited investigation into whether these conclusions generalize to the diverse Vision Transformer (ViT) architectures that now dominate computer vision. This paper presents a controlled benchmark that evaluates attribution quality across five dimensions: faithfulness, localization, robustness, complexity, and computational cost. A standardized framework assesses 13 attribution methods from four algorithmic families on eight representative backbones spanning CNNs, isotropic ViTs, hierarchical transformers, hybrid architectures, and linear-attention transformers. The results show that attribution performance is strongly architecture-dependent and that rankings established on CNNs do not reliably transfer to transformer-based models. CAM-based methods achieve the highest scores under the conventional bounding-box localization metric on CNNs and most ViTs but perform poorly on linear-attention architectures. Pixel-level dense-mask evaluation further reveals that these gains largely reflect metric saturation rather than accurate localization. CAM-based methods also exhibit limited robustness on global-attention transformers, whereas attention rollout provides consistently stable explanations with poor localization. Furthermore, faithfulness correlation offers limited discrimination between attribution methods, highlighting the limitations of single-metric evaluation. These findings challenge prevailing conclusions on attribution performance and demonstrate the need for architecture-aware, multi-dimensional evaluation. The open-source code for the evaluation framework and benchmark results is available at \url{https://github.com/Nishan-Charlie/VIT_XAI_Bench}.
\end{abstract}


\end{frontmatter}

\section{Introduction}
\label{sec:introduction}

Deep learning (DL) has significantly advanced modern computer vision through improved performance in image classification, object detection, semantic segmentation, and visual recognition~\cite{russakovsky2015imagenet, he2016resnet, liu2022convnext}. The improved predictive capability of deep neural networks has expanded their application across scientific, industrial, and safety-critical domains. In these high-stakes settings, predictive accuracy alone is insufficient, as model decisions often require inspection, verification, and technical justification~\cite{nauta2023anecdotal}. This need has accelerated research into explainable artificial intelligence (XAI) methods, particularly post-hoc attribution techniques that identify the visual evidence associated with model predictions~\cite{bach2015lrp, bodria2023benchmarking}.

Over the past decade, Convolutional Neural Networks (CNNs) and Vision Transformers (ViTs) have emerged as the two dominant architectures for visual representation learning. CNNs construct representations through local receptive fields and hierarchical spatial feature extraction, incorporating strong local inductive biases~\cite{he2016resnet}. In contrast, ViTs partition images into discrete patches, represent them as tokens, and capture global dependencies through multi-head self-attention (MHSA) mechanisms~\cite{dosovitskiy2021vit}. These architectures process and propagate visual information through fundamentally different internal structures~\cite{raghu2021vit}. Consequently, post-hoc XAI methods exhibit architecture-dependent behavior across different backbones~\cite{chefer2021transformer, wu2024faithfulness}. Existing attribution methods commonly rely on backward gradients~\cite{simonyan2014saliency, sundararajan2017ig}, intermediate activation maps~\cite{selvaraju2017gradcam}, or forward input perturbations~\cite{zeiler2014occlusion, petsiuk2018rise}. A method that provides reliable explanations for CNNs may therefore produce inconsistent or unreliable results when applied to ViTs~\cite{chefer2021transformer}. This limitation arises because many XAI methods assume continuous spatial representations, whereas the discrete tokenization and attention-based information flow in ViTs introduce different attribution characteristics~\cite{abnar2020rollout, achtibat2024attnlrp}.

Despite the importance of this problem, existing XAI evaluations remain fragmented, with variations in data pipelines, attribution implementations, and metric configurations~\cite{bodria2023benchmarking, brandt2023precise}. This lack of standardization restricts direct comparison and provides limited evidence on the transferability of attribution methods across visual architectures. Moreover, attribution evaluation involves multiple objectives, where each metric captures distinct properties of explanations, including faithfulness, localization, robustness, and computational cost~\cite{nauta2023anecdotal, hedstrom2023quantus}. Since strong performance on one metric does not necessarily translate to others, reliable comparison requires a unified framework with consistent experimental settings.

To address these limitations, this study presents a controlled benchmark to investigate whether attribution behavior transfers consistently across CNN and ViT architectures. A standardized evaluation framework assesses 13 attribution methods from four algorithmic families: gradient-based (Saliency, IG, Input$\times$Gradient, SmoothGrad, VarGrad, GradientSHAP), CAM-based (Grad-CAM, Grad-CAM++), attention-based (Attention Rollout, AttnLRP), and perturbation-based (Occlusion, RISE, LIME), as summarized in Table~\ref{tab:method_summary}. All methods are evaluated on the same eight backbones, spanning the evolution of ViTs from isotropic models~\cite{dosovitskiy2021vit, touvron2021deit} to hierarchical, hybrid, and linear-attention architectures~\cite{liu2021swin, mehta2022mobilevit, cai2023efficientvit}. The attribution method families and evaluation axes represent two orthogonal dimensions of the benchmark. The former categorizes explanation mechanisms, whereas the latter assesses explanation quality across five complementary properties.

\begin{table}[ht]
\centering
\caption{Summary of benchmarked post-hoc attribution methods and their applicability to ViT and CNN. A green check mark (\cmark) indicates that the method is structurally applicable to the architecture family and is included in the corresponding benchmark cells; a red cross (\xmark) indicates that the method cannot be computed for that family (attention-native methods require attention matrices, which CNNs do not possess). A parenthesized check (\cmark)$^{*}$ marks partial applicability: Attention Rollout and AttnLRP apply only to the isotropic ViT-B/16, which exposes a global class-token attention matrix, and are undefined on the hierarchical, multi-axis, hybrid, and linear-attention backbones, i.e.\ on six of the seven ViT variants evaluated (Section~\ref{sec:framework}).}

\label{tab:method_summary}
\begin{tabular}{llcc}
\toprule
\textbf{Method} & \textbf{Family} & \textbf{ViT} & \textbf{CNN} \\
\midrule
Saliency \cite{simonyan2014saliency} & Gradient & \cmark & \cmark \\
IG \cite{sundararajan2017ig} & Gradient & \cmark & \cmark \\
Input$\times$Gradient \cite{shrikumar2017deeplift} & Gradient & \cmark & \cmark \\
SmoothGrad \cite{smilkov2017smoothgrad} & Gradient & \cmark & \cmark \\
VarGrad \cite{adebayo2018sanity} & Gradient & \cmark & \cmark \\
GradientSHAP \cite{lundberg2017shap} & Gradient & \cmark & \cmark \\
\midrule
Grad-CAM \cite{selvaraju2017gradcam} & CAM & \cmark & \cmark \\
Grad-CAM++ \cite{chattopadhay2018gradcampp} & CAM & \cmark & \cmark \\
\midrule
Attention Rollout \cite{abnar2020rollout} & Attention & (\cmark)$^{*}$ & \xmark \\
AttnLRP \cite{achtibat2024attnlrp} & Attention & (\cmark)$^{*}$ & \xmark \\
\midrule
Occlusion \cite{zeiler2014occlusion} & Perturbation & \cmark & \cmark \\
RISE \cite{petsiuk2018rise} & Perturbation & \cmark & \cmark \\
LIME \cite{ribeiro2016lime} & Perturbation & \cmark & \cmark \\
\bottomrule
\end{tabular}
\end{table}

In contrast to existing XAI benchmarks that focus on a single model family~\cite{wu2024faithfulness}, that use synthetic toy datasets~\cite{hesse2023funnybirds, arras2022clevrxai}, or that evaluate individual metrics~\cite{hooker2019roar, rao2022attribution}, this work isolates the impact of architectural differences on explanation behavior. The key contributions are as follows:

\begin{itemize}[leftmargin=*]
    \item \textbf{Controlled Cross-Architecture Benchmark:} We establish a unified evaluation framework that controls data, target selection, preprocessing, and experimental settings to enable a systematic comparison of attribution methods across  CNNs, isotropic ViTs, and advanced hierarchical/efficient transformer backbones.
    
    \item \textbf{Multi-Axis Evaluation Suite:} We evaluate 13 attribution methods from four algorithmic families across five dimensions of explanation quality: \textit{faithfulness} using Faithfulness Correlation (FC)~\cite{bhatt2020faithfulness} and Faithfulness Estimate (FE)~\cite{alvarez2018robustness}, \textit{localization} using the Pointing Game and a dense-mask Energy-Based Pointing Game (PG, EBPG)~\cite{zhang2018pointing}, \textit{robustness} using Max-Sensitivity~(MS) \cite{yeh2019sensitivity}, \textit{complexity} using Sparseness (SP)~\cite{chalasani2020sparseness} and computational cost using wall-clock time per explanation. The analysis demonstrates that attribution rankings are architecture-dependent, with methods achieving strong performance on CNNs often exhibiting substantial degradation or rank reversal on transformer-based models.
    
    \item \textbf{Architectural Vulnerability Analysis:} We provide theoretical and empirical analyses of the influence of architectural design on attribution reliability across CNNs and transformer-based models. The study identifies architecture-specific failure modes, including the collapse of Grad-CAM++~\cite{chattopadhay2018gradcampp} under global self-attention in ViTs~\cite{dosovitskiy2021vit} and the degradation of gradient-based methods caused by local window partitioning in Swin Transformers~\cite{liu2021swin}.
    
    \item \textbf{Open-Source XAI Evaluation Framework\footnote{The harness, configuration files, and raw per-image outputs will be made publicly available upon acceptance.}:} We release an extensible registry-based benchmarking framework and comprehensive empirical results, including configuration files and per-sample evaluation outputs, to facilitate reproducible and scalable cross-architecture XAI research.
\end{itemize}

The remainder of this paper is organized as follows. Section~\ref{sec:related} reviews related work on attribution methods, evaluation strategies, and the gaps in cross-architecture benchmarking. Section~\ref{sec:vit_evolution} outlines the evolution of Vision Transformers and the associated architectural challenges. Section~\ref{sec:framework} describes the proposed benchmark framework, datasets, and evaluation metrics, while Section~\ref{sec:setup} presents the experimental setup. Section~\ref{sec:results} reports the quantitative and qualitative results, and Section~\ref{sec:discussion} discusses architectural failure modes and attribution transferability. Finally, Section~\ref{sec:conclusion} concludes the paper and outlines future research directions.

\section{Related Work}
\label{sec:related}

\subsection{Architectural Divergence and Attribution Mechanics}
The initial wave of post-hoc explainability was heavily optimized for CNNs~\cite{zeiler2014occlusion,bach2015lrp}. Foundational gradient methods such as Saliency~\cite{simonyan2014saliency}, IG~\cite{sundararajan2017ig}, and SmoothGrad~\cite{smilkov2017smoothgrad} were designed to exploit the continuous and hierarchical gradient flow of stacked convolutions. Similarly, Class Activation Mapping (CAM) and its variants, such as Grad-CAM~\cite{selvaraju2017gradcam}, were explicitly engineered under the assumption that spatial feature maps retain a linear relationship with the final classification logit through Global Average Pooling (GAP). 

When ViTs emerged, researchers quickly realized that legacy methods tailored for CNNs often produced noisy or uninterpretable results on self-attention architectures~\cite{chefer2021transformer}. The introduction of discrete token graphs, nonlinear softmax bottlenecks, and centralized classification tokens broke the structural assumptions of gradient and CAM techniques. This architectural shift prompted the development of attention-specific XAI techniques. Methods like Attention Rollout~\cite{abnar2020rollout} attempt to trace information flow through the discrete token graph, while specialized approaches like AttnLRP~\cite{achtibat2024attnlrp} modify Deep Taylor Decomposition to navigate the softmax operations inherent to transformer models. However, the literature has largely treated CNN and ViT explainability as isolated silos. Novel explanation methods are routinely proposed for one specific architecture but are rarely stress-tested across the structural divide to verify their general applicability.

\subsection{The Fragmentation of Quantitative XAI Evaluation}
Because human-independent ground truth for visual explanations is rarely available outside of synthetic environments, the community has shifted toward axiomatic and surrogate evaluation criteria. Evaluating explanation quality is now widely understood as a multi-objective problem~\cite{nauta2023anecdotal,bhatt2020faithfulness}. Faithfulness metrics measure whether the attribution scores accurately reflect the internal reasoning of the model by tracking output degradation during feature occlusion. Localization metrics, such as the PG~\cite{zhang2018pointing}, assess spatial alignment against bounding boxes or semantic masks. Robustness metrics quantify the stability of an explanation against adversarial or imperceptible input perturbations, while complexity metrics evaluate the spatial concentration of the resulting heatmap.

To standardize these disparate evaluation axes, unified diagnostic libraries such as Quantus~\cite{hedstrom2023quantus} have been developed. Yet, despite the availability of standardized metric implementations, researchers frequently report conflicting findings~\cite{rao2022attribution,brandt2023precise}. These discrepancies arise because fundamental evaluation parameters such as perturbation baselines, dataset complexity, and masking thresholds are inconsistently applied across different studies. This methodological fragmentation makes it impossible to reliably compare an attribution method evaluated in one paper against a competing method evaluated in another. By anchoring our evaluation framework in a rigidly parameterized Quantus environment, we eliminate these confounding variables, ensuring that performance variations reflect genuine architectural differences rather than arbitrary metric configurations.

\subsection{The Gap in Cross-Architecture Benchmarking}

\begin{table*}[t]
\centering
\caption{Comparison with prior quantitative XAI evaluation studies. Method families: G = gradient-based, C = CAM-based, A = attention-native, and P = perturbation-based. Evaluation axes: F = faithfulness, L = localization, R = robustness, S = sparseness/complexity, and T = computational cost. Modern ViT covers hierarchical, multi-axis, hybrid, and linear-attention backbones.}
\label{tab:benchmark_comparison}
\setlength{\tabcolsep}{5pt}
\begin{tabular}{lccccccc}
\toprule
\textbf{Benchmark} & \textbf{CNN} & \textbf{Iso.\ ViT} & \textbf{Modern ViT} & \textbf{Real Data} & \textbf{Families} & \textbf{Axes} & \textbf{Controlled cross-arch.} \\
\midrule
ROAR~\cite{hooker2019roar} & \cmark & \xmark & \xmark & \cmark & G & F & \xmark \\
CLEVR-XAI~\cite{arras2022clevrxai} & \cmark & \xmark & \xmark & \xmark & G, C & L & \xmark \\
Rao et al.~\cite{rao2022attribution} & \cmark & \xmark & \xmark & \cmark & G, C & L & \xmark \\
Bodria et al.~\cite{bodria2023benchmarking} & \cmark & \xmark & \xmark & \cmark & G, P & F, R & \xmark \\
FunnyBirds~\cite{hesse2023funnybirds} & \cmark & \cmark & \xmark & \xmark & G, C, P & F, L & \cmark \\
Wu et al.~\cite{wu2024faithfulness} & \xmark & \cmark & \xmark & \cmark & G, A & F & \xmark \\
\midrule
\textbf{Ours} & \cmark & \cmark & \cmark & \cmark & G, C, A, P & F, L, R, S, T & \cmark \\
\bottomrule
\end{tabular}
\end{table*}

Most existing large-scale XAI benchmarks are fundamentally constrained by their reliance on legacy CNN backbones such as ResNet and VGG~\cite{hooker2019roar,rao2022attribution,brandt2023precise}. While recent works have begun to investigate the faithfulness of ViT explanations in isolation~\cite{wu2024faithfulness}, they generally do not provide a strictly controlled baseline against convolutional models. Furthermore, existing evaluation frameworks tend to treat the Vision Transformer as a single and static architecture. The current XAI benchmarking literature largely ignores the rapid evolutionary branching of vision models into hierarchical designs like Swin~\cite{liu2021swin}, multi-axis frameworks like MaxViT~\cite{tu2022maxvit}, and linear attention architectures like EfficientViT~\cite{cai2023efficientvit}. 

To our knowledge, no prior work performs a rigidly controlled comparison across this evolutionary timeline. The literature currently lacks a unified empirical study where training data, computational budgets, and metric implementations are held strictly constant to isolate the specific impact of architectural bottlenecks on XAI transferability. Table~\ref{tab:benchmark_comparison} positions our study against the most closely related quantitative evaluation efforts and makes this coverage gap explicit: no prior benchmark spans all four attribution families and all five quality axes across CNN, isotropic ViT, and modern ViT backbones on real data under a single fixed protocol. By subjecting these modern structural variants to the exact same diagnostic gauntlet as standard ViTs and CNNs, this benchmark isolates precisely where and why explanation transferability breaks down.

\section{Evolution of the Vision Transformer Family and XAI Bottlenecks}
\label{sec:vit_evolution}

The paradigm shift from CNNs to ViTs fundamentally altered representation learning. While CNNs enforce a strong inductive bias through local receptive fields and translation equivariance, Transformers treat an image as a sequence of discrete tokens processed by global self-attention. This section surveys the chronological evolution of the ViT family from the original isotropic models to modern, highly efficient architectures. Tracing this evolution is critical because the architectural divergence from the original ViT is precisely what breaks standard post-hoc XAI methods.

\subsection{The Genesis: The Original Vision Transformer (2020)}
The original ViT, introduced by Dosovitskiy et~al.~\cite{dosovitskiy2021vit}, demonstrated that a pure transformer architecture could achieve state-of-the-art image classification without relying on convolutional downsampling. The architecture splits an image into non-overlapping patches, projects them linearly into embeddings, prepends a learnable class token (\texttt{[CLS]}), and processes the sequence via standard MHSA. For XAI, the introduction of this \texttt{[CLS]} token and the softmax attention bottleneck fundamentally violated the spatial linearity assumptions of CNN-based methods like Grad-CAM~\cite{chefer2021transformer}.

\subsection{Phase 1: Training Data Efficiency (Early 2021)}
To resolve the extreme data hunger of the original ViT, the Data-efficient Image Transformer (DeiT)~\cite{touvron2021deit} was introduced. DeiT proved that ViTs could be trained on the standard ImageNet-1k~\cite{russakovsky2015imagenet} dataset using a novel token-based knowledge distillation strategy, effectively embedding the inductive biases of a strong convolutional teacher into a pure transformer architecture. 

\subsection{Phase 2: Hierarchical Structures and Inductive Biases (Late 2021)}
The isotropic structure of the original ViT scaled poorly to dense prediction tasks requiring multi-scale feature maps. This motivated the reintroduction of local inductive biases. The Swin Transformer~\cite{liu2021swin} replaced global attention with shifted-window self-attention, achieving linear computational complexity. The Pyramid Vision Transformer (PVT)~\cite{wang2021pvt} took a complementary route, building a multi-stage feature pyramid and taming the cost of global attention through spatial-reduction attention, which downsamples the keys and values. MaxViT~\cite{tu2022maxvit} expanded on this by computing attention across both a local block and a sparse, global grid. For gradient-based XAI, these localized windows constrain gradient routing between tokens in different windows, which can leave window-aligned artifacts in the resulting maps.

\subsection{Phase 3: Hybrid and Efficient Refinements (2022--Present)}
As the ViT family branched into extreme deployment scenarios, structural homogeneity disappeared. At the edge, MobileViT~\cite{mehta2022mobilevit} replaced standard transformers with a lightweight convolutional-transformer hybrid design. Concurrently, EfficientViT~\cite{cai2023efficientvit} substituted standard softmax attention with an efficient, hardware-friendly linear formulation. The transition from global softmax attention to convolutions and linear attention changes gradient routing, which can cause legacy XAI methods to produce noisy or collapsed attribution maps on some of these backbones.

\section{Post-Hoc Attribution Methods in ViTs}
\label{sec:attribution_methods}

\begin{figure}
    \centering
    \includegraphics[width=\linewidth]{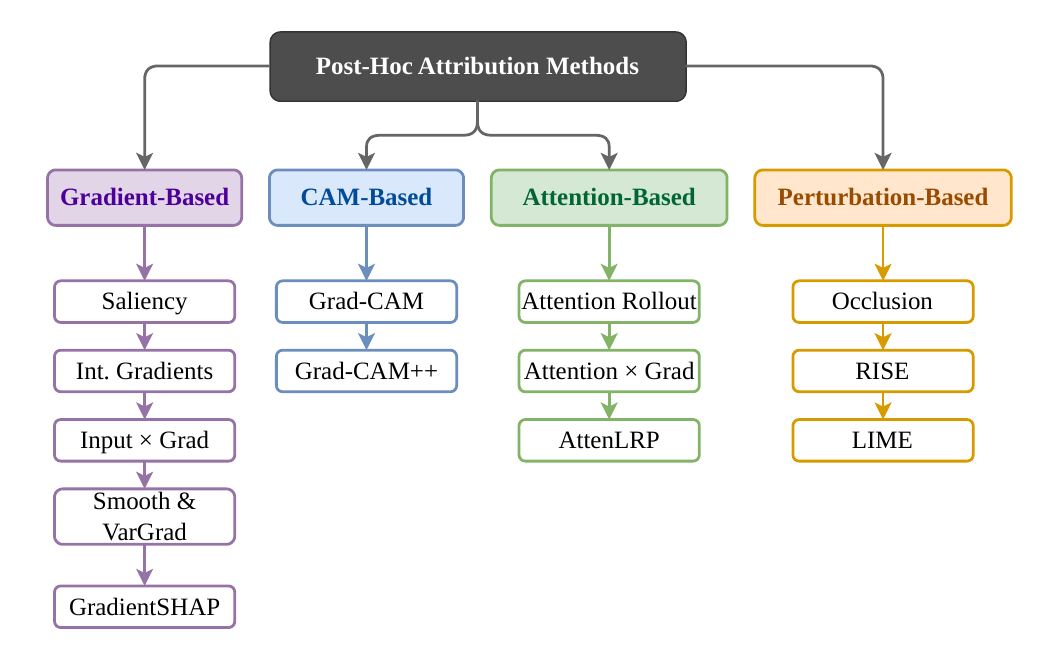}
    \caption{Taxonomy of the four post-hoc attribution families evaluated in this benchmark, together with the structural assumption each family places on the backbone. These assumptions are exactly what modern ViT variants break.}
    \label{fig:method_family}
\end{figure}

While ViTs have achieved remarkable predictive performance, their internal mechanisms, which rely on self-attention and patch-based tokenization rather than local convolutions, pose unique challenges for interpretability. Post-hoc attribution methods attempt to map the model's decision back to the input space, assigning a relevance score to each pixel or patch.

In this benchmark, we evaluate 13 distinct attribution methods categorized into four primary families (Figure~\ref{fig:method_family}): Gradient-based, CAM-based, Attention-native, and Perturbation-based. Table~\ref{tab:method_summary} summarizes these methods and their architectural compatibility.

\subsection{Gradient-Based Approaches}
Gradient-based methods rely on backpropagating the classification score of a target class $c$, denoted as $F_c(x)$, to the input space $x$. 

\textbf{Saliency (Vanilla Gradients):} The foundational approach defines the attribution map $M$ as the absolute value of the gradient of the output with respect to the input:
\begin{equation}
M_{Sal}(x) = \left| \frac{\partial F_c(x)}{\partial x} \right|
\end{equation}

\textbf{IG:} IG satisfies the axioms of sensitivity and implementation invariance by accumulating gradients along a straight-line path from a baseline image $x'$ to the input $x$:
\begin{equation}
M_{IG}(x) = (x - x') \times \int_{\alpha=0}^{1} \frac{\partial F_c(x' + \alpha(x - x'))}{\partial x} d\alpha
\end{equation}

\textbf{SmoothGrad \& VarGrad:} Because gradients in deep architectures can be noisy, SmoothGrad computes the expectation of gradients over inputs perturbed by Gaussian noise $\mathcal{N}(0, \sigma^2)$:
\begin{equation}
M_{SG}(x) = \frac{1}{N} \sum_{i=1}^{N} \frac{\partial F_c(x + \mathcal{N}(0, \sigma^2))}{\partial x}
\end{equation}
where $N$ denotes the number of sampled noise instances. VarGrad acts as a natural extension of this concept, computing the variance of these perturbed gradients rather than their mean, which further isolates robust structural features.

\subsection{CAM-Based Adaptations}
For ViTs, CAM methods are adapted by utilizing the spatial token embeddings (excluding the \texttt{[CLS]} token) from the final encoder block. Grad-CAM computes a weighted sum of the feature maps $A^k$:
\begin{equation}
M_{GC}(x)=\operatorname{ReLU}\!\left(\sum_{k}\alpha_k^c A^k\right)
\end{equation}
where the weights $\alpha_k^c$ are derived from the global average of the gradients:
\begin{equation}
\alpha_k^c=\frac{1}{Z}\sum_{i}\sum_{j}
\frac{\partial F_c}{\partial A_{i,j}^k}
\end{equation}
Here $A^k$ denotes the $k$-th spatial feature map of the selected layer (for ViTs, the spatial token embeddings reshaped to a 2-D grid), $A^k_{i,j}$ is its activation at location $(i,j)$, $\alpha_k^c$ is the importance weight of feature map $k$ for the target class $c$, and $Z$ is the number of spatial locations in the feature map, so that the sum implements global average pooling of the gradients.

\subsection{Attention-Native Methods}
Unlike CNNs, ViTs possess a native routing mechanism: self-attention. These methods directly exploit the attention matrices to trace information flow and token interactions.

\textbf{Attention Rollout:} This method models the flow of information across layers as a Markov chain, adding an identity matrix $I$ to the raw attention matrix $A_L$ at layer $L$ to account for residual connections:
\begin{equation}
R_L = \frac{1}{2}(A_L + I) R_{L-1}
\end{equation}
where $R_L$ is the accumulated token-to-token rollout matrix after layer $L$, $R_{L-1}$ is the rollout of the preceding layer with the recursion initialized at the identity, $R_0 = I$, and $A_L$ is the head-averaged attention matrix of layer $L$. The factor $\tfrac{1}{2}$ renormalizes the equal mixture of the attention and residual (identity) paths. The final map is read from the row of $R_L$ associated with the classification token.

\begin{figure*}[t]
    \centering
    \includegraphics[width=0.92\linewidth]{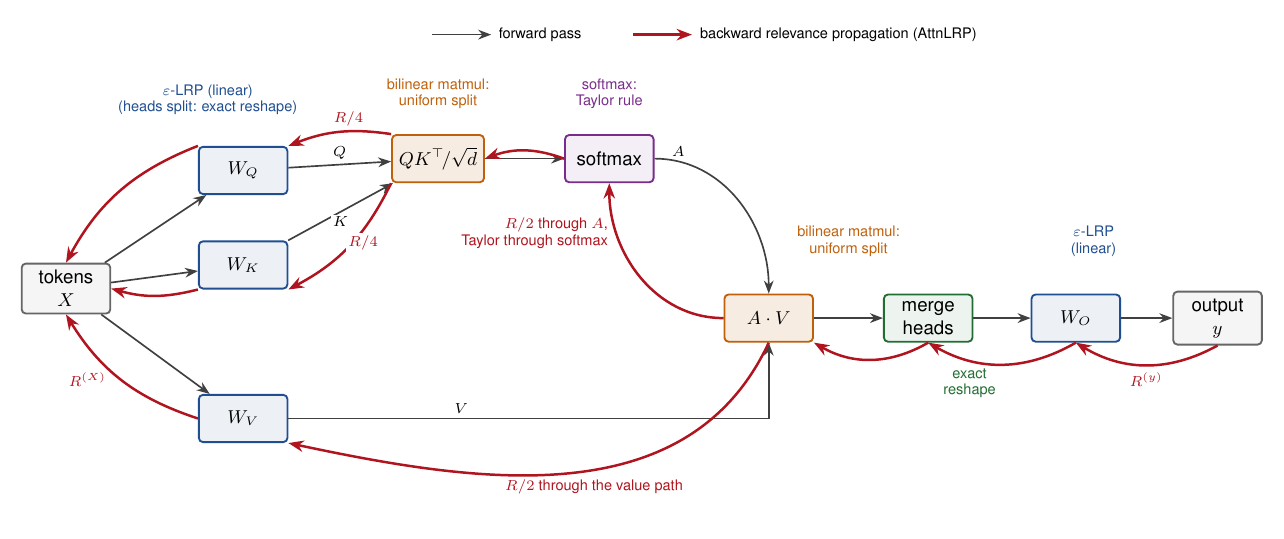}
    \caption{AttnLRP relevance propagation through a multi-head attention
  block~\cite{achtibat2024attnlrp}. The forward computation (black) is left
  unchanged; the backward pass (red) redefines the relevance flow per operation:
  $\varepsilon$-LRP through the linear projections, a uniform bilinear split at
  the two matmuls (half the relevance to each operand), a Taylor rule through
  the softmax, and exact reshapes for head splitting and merging.}
    \label{fig:attn_lrp}
\end{figure*}

\textbf{AttnLRP (Transformer-Specific LRP):} Standard Layer-wise Relevance Propagation (LRP) rules often fail when applied to ViTs due to complex token interactions. AttnLRP specifically tailors the Deep Taylor Decomposition framework to handle the MHSA module (Figure~\ref{fig:attn_lrp}).

In the MHSA module, relevance is propagated backward through two bilinear products and one softmax. The attention output $\mathbf{O} = \mathbf{A}\mathbf{V}$ is bilinear in the attention matrix $\mathbf{A}$ and the values $\mathbf{V}$, so the incoming relevance $R_{\mathbf{O}}$ is split between the two factors by an $\varepsilon$-stabilized bilinear rule that preserves the total relevance:
\begin{equation}
\textstyle\sum R_{\mathbf{A}} + \sum R_{\mathbf{V}} = \sum R_{\mathbf{O}}.
\end{equation}
The softmax itself is handled by a first-order Taylor expansion at the evaluation point, which renders it locally linear and redistributes $R_{\mathbf{A}}$ onto the pre-softmax logits $\mathbf{Z} = \mathbf{Q}\mathbf{K}^\top/\sqrt{d}$:
\begin{equation}
R_{Z_{ij}} = Z_{ij} \sum_{k} \frac{\partial A_{ik}}{\partial Z_{ij}} \, \frac{R_{A_{ik}}}{A_{ik}}.
\end{equation}
Here $Z_{ij}$ is the pre-softmax attention logit between query token $i$ and key token $j$, $A_{ik}$ is the corresponding post-softmax attention weight, $R_{A_{ik}}$ is the relevance assigned to that attention entry by the bilinear rule above, and $R_{Z_{ij}}$ is the relevance redistributed onto the logit. Because $\mathbf{Z}$ is itself bilinear in the queries $\mathbf{Q}$ and keys $\mathbf{K}$, the same bilinear rule finally splits the logit relevance $R_{\mathbf{Z}}$ into the query relevance $R_{\mathbf{Q}}$ and the key relevance $R_{\mathbf{K}}$, completing the backward pass through the attention head.

\subsection{Perturbation-Based Methods}
These methods treat the underlying architecture as a black box, measuring the change in the output probability when parts of the input are altered.

\textbf{RISE:} RISE generates thousands of random binary masks $M_i$ and passes the masked images through the model. The final attribution map is the expected value of the masks weighted by the output score:
\begin{equation}
M_{RISE}(x) = \frac{1}{N \, \mathbb{E}[M]} \sum_{i=1}^{N} F_c(x \odot M_i) \cdot M_i
\end{equation}
where $M_i \in \{0,1\}^{H \times W}$ is the $i$-th random binary occlusion mask (upsampled to input resolution), $N$ is the number of sampled masks, $\odot$ denotes elementwise multiplication, and the normalization by the expected mask value $\mathbb{E}[M]$ corrects for the frequency with which each pixel is left visible across the mask ensemble.

\textbf{Occlusion:} A foundational perturbation technique that systematically slides a gray or zero-value mask across the input image, mapping the drop in target class probability to the occluded region.

\textbf{LIME:} LIME trains an inherently interpretable surrogate model (typically a linear ridge regression) on a local neighborhood of perturbed super-pixels to approximate the complex model's behavior in the immediate vicinity of the target input.

\section{Proposed Benchmark Framework}
\label{sec:framework}

\begin{figure*}[t]
  \centering
  \includegraphics[width=0.97\textwidth]{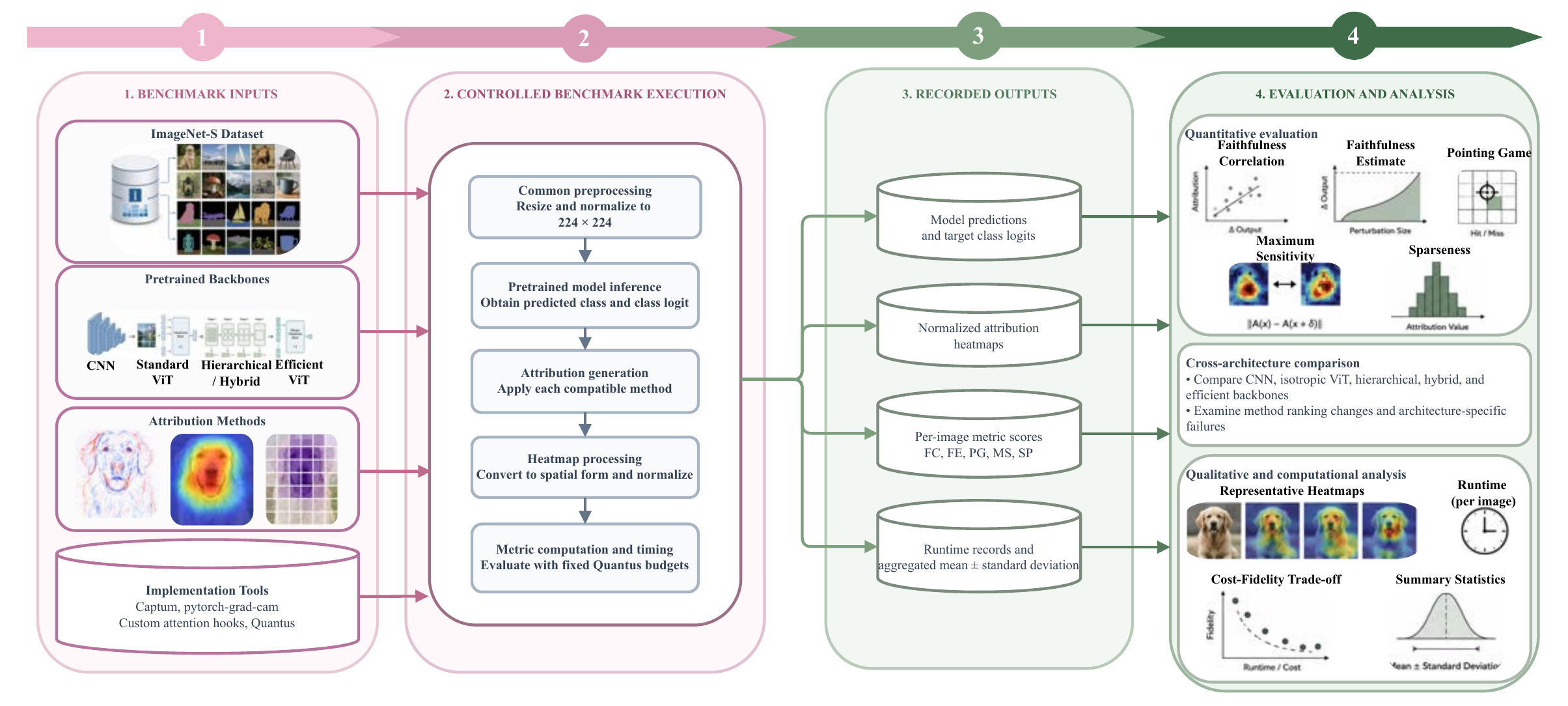}
  \caption{Overview of the benchmark framework. A fixed ImageNet-S image set is explained by 13 attribution methods on 8 backbones; all heatmaps pass through one shared normalization protocol, and five \textsc{Quantus} metrics score faithfulness, localization, robustness, and complexity. The color chips mark the four method families used throughout the paper.}
  \label{fig:pipeline}
\end{figure*}

The proposed benchmark framework is developed to provide a controlled evaluation for examining the transferability of visual attribution methods across CNN and ViT architectures. The framework organizes the evaluation around a fixed grid of models, attribution methods, datasets, and metrics. Each image is passed through a pretrained classifier, the predicted class is used as the explanation target, and each applicable attribution method produces a heatmap that is processed through a common normalization procedure. The same data, target selection rule, heatmap processing procedure, and metric configuration are applied across all model and method combinations. Figure~\ref{fig:pipeline} summarizes the pipeline.

\subsection{Datasets}\label{sec:dataset}
The benchmark is conducted using the validation split of ImageNet-S~\cite{gao2022imagenets}, an extension of ImageNet-1k~\cite{russakovsky2015imagenet} that provides dense semantic segmentation masks for corresponding images. For localization scoring, we reduce each dense mask to its tight axis-aligned bounding box, so the Pointing Game target is the rectangular region that encloses the object rather than its exact silhouette. A fixed subset of 1,000 validation images is used across all model and attribution method combinations to remove sampling variation and strictly isolate architectural factors.

\subsection{Models Evaluated}\label{sec:model}
The benchmark's model space deliberately spans architecture design, not merely scale. Beyond the plain isotropic ViT, modern practice is dominated by efficient and hierarchical transformers that replace global self-attention with structurally different mechanisms. We therefore evaluate backbones drawn from five advanced families: Swin~\cite{liu2021swin} (shifted-window attention), PVT-v2~\cite{wang2021pvt} (spatial-reduction attention), MaxViT~\cite{tu2022maxvit} (multi-axis attention), MobileViT~\cite{mehta2022mobilevit} (convolution-transformer hybrid), and EfficientViT~\cite{cai2023efficientvit} (multi-scale linear attention; abbreviated EffViT in the tables).

Crucially, these backbones are not flat token-sequence models: none carries a global class token, and their last-stage features are spatial maps rather than a patch grid. Across the metric tables of Section~\ref{sec:results}, ViT-B/16 and ResNet-50 serve as the isotropic-ViT and CNN reference points, while Swin-B, PVT-v2-B2, MaxViT-S, MobileViT-v2, and EfficientViT-B1/B2 expose the limits of current XAI methods on hierarchical, multi-axis, hybrid, and linear-attention designs. These backbones are matched by tier rather than by exact parameter count (they range from roughly $9$M to $88$M parameters), so this design does not separate architecture from model capacity, and we do not claim that it does. We therefore restrict our conclusions to the \emph{direction} of failure rather than its exact magnitude. We anchor each to a specific operation through the mechanism analyses of Sections~\ref{sec:quant}, \ref{sec:gradcam_collapse}, and \ref{sec:swin_bottlenecks}: linear attention collapses CAM localization at both EfficientViT scales (Grad-CAM Pointing Game $0.70$ at $9$M and $0.55$ at $24$M, both far below the CNN and, under the dense metric, near the random floor), Grad-CAM++ destabilizes under global attention across sizes, and the coarse $7\times7$ terminal grid appears at every hierarchical scale. The residual quantitative differences, such as the $0.70$-versus-$0.55$ gap between the two EfficientViT sizes, may well include a capacity component that this design cannot isolate; disentangling architecture from scale with an iso-parameter sweep is left to future work.

\subsection{Attribution Methods}
The benchmark evaluates 13 attribution methods grouped into four families, as previously detailed in Table~\ref{tab:method_summary}. Gradient-based, CAM-based, and perturbation-based methods are largely architecture-agnostic (with CAM requiring a token-to-grid reshape for ViTs).

Attention-native methods are treated separately because their applicability depends on the availability of suitable attention representations, so neither method is applied to the CNN. Attention Rollout reads a global token-to-token attention matrix from a class token. Only the isotropic ViT-B/16 exposes such a matrix, so Rollout is reported for ViT-B/16 alone and is left undefined on the hierarchical, multi-axis, hybrid, and linear-attention backbones, none of which carry a global class token.

AttnLRP is reported only for the isotropic ViT-B/16. On that backbone, we use the published recipe of Achtibat et~al.~\cite{achtibat2024attnlrp}: the softmax Taylor rule and the uniform bilinear split for standard multi-head self-attention. The released toolbox provides no bespoke rules for windowed (Swin), spatial-reduction (PVT-v2), multi-axis (MaxViT), convolution-hybrid (MobileViT), or linear cross-covariance (EfficientViT) attention; on those backbones it can only fall back to a generic composite with no attention-specific decomposition. Reporting such fallback numbers would confound a genuine transfer failure with the artifact of running an unmodified, incompatible algorithm, so we leave those cells undefined rather than present them as AttnLRP results. This gap is itself part of the transferability finding: the current attention-native toolchain has no defined rules for the attention mechanisms that now dominate the ViT family. Standard Captum LRP is even more restrictive and cannot be executed on these backbones, as it rejects unsupported modules in each family (Section~\ref{sec:attribution_methods}).

\subsection{Evaluation Metrics}
\begin{figure}[!t]
  \centering
  \includegraphics[width=0.96\columnwidth]{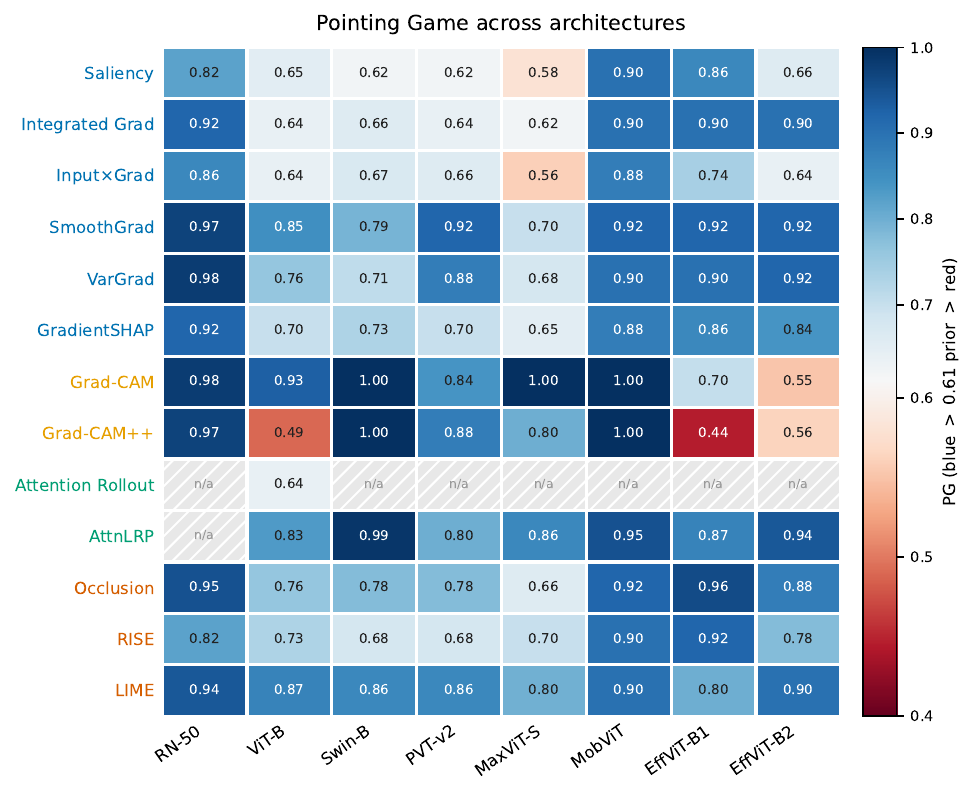}
  \caption{\textbf{Localization breaks by architecture, not uniformly.} PG across 13 methods and 8 backbones, colored diverging around the $0.61$ random-point prior (blue $=$ localizes, red $=$ below chance; hatched $=$ not applicable). 
  }
  \label{fig:pointing}
\end{figure}

The benchmark evaluates attribution quality using five metrics from four \textsc{Quantus}~\cite{hedstrom2023quantus} categories. FC and FE measure the agreement between attribution values and output changes caused by feature modification. Localization is evaluated using the PG: for each image, we take the tight axis-aligned bounding box of the ImageNet-S mask and record a hit when the single maximum-attribution pixel lands inside that box. The PG is therefore a soft, best-case localization test. It inspects only the peak pixel against a filled rectangle, not the full heatmap against the object silhouette. It also carries a high floor that we can derive rather than assert: because the target is a large, typically central box, a uniformly random peak already lands inside it with probability equal to the mean bounding-box area fraction, which is $0.59$ on this set, close to the empirical $0.61$ random-point baseline in Figure~\ref{fig:pointing}. We therefore additionally report the Energy-Based Pointing Game (EBPG), the fraction of attribution energy that falls inside the pixel-perfect segmentation mask; EBPG does not saturate and is our localization metric of record (Table~\ref{tab:ebpg}). MS measures the stability of an attribution map under small input changes, while SP measures the concentration of attribution values using a Gini-style coefficient. All metrics are computed using \textsc{Quantus} with standardized sampling budgets.

\FloatBarrier
\section{Experimental Setup}
\label{sec:setup}

\subsection{Implementation Details}
All experiments were conducted at a standard $224 \times 224$ input resolution using PyTorch~$2.6$ on a single NVIDIA GPU. Images are resized to $224 \times 224$ and normalized with the standard ImageNet statistics (mean $[0.485, 0.456, 0.406]$, standard deviation $[0.229, 0.224, 0.225]$), and the predicted class is used as the explanation target. Attribution methods use standardized community libraries: gradient and perturbation methods via \texttt{captum}~\cite{kokhlikyan2020captum}, CAM variants via \texttt{pytorch-grad-cam}~\cite{gildenblat2021pytorchgradcam}, and attention-native methods via custom forward hooks; all metrics use \textsc{Quantus}~$0.6$~\cite{hedstrom2023quantus}, and all backbones are the public \texttt{timm}~$1.0$ checkpoints. For CAM, the target layer is the terminal feature map returned by each backbone's \texttt{forward\_features} (a $7 \times 7$ grid for the hierarchical, multi-axis, and linear-attention backbones); for the isotropic ViT, we target the input normalization of the last transformer block, because its head pools only the class token and the terminal token grid receives no gradient. This per-architecture target is forced by the models rather than tuned: the terminal feature map is the natural CAM target for the CNN and the hierarchical, hybrid, and linear backbones, and the last-block normalization is the standard \texttt{pytorch-grad-cam} recipe for the isotropic ViT. We do not search over layers, so the CAM comparison reflects each backbone's canonical configuration, not a per-model tuning. Stochastic methods (SmoothGrad, VarGrad, RISE, and the internal \textsc{Quantus} perturbations) use a fixed seed of $0$ for the reported single-seed run.

\subsection{Attribution Normalization}
Because attribution methods yield outputs on vastly different numerical scales, and because gradient methods produce signed values while CAM outputs are strictly positive, all heatmaps undergo a strict normalization protocol prior to metric evaluation. For each generated attribution map, we take the absolute value of the relevance scores and apply min-max scaling, mapping all heatmap values to a strict $[0, 1]$ range. This ensures that the evaluation metrics assess the spatial distribution of relevance rather than arbitrary scale differences.

\subsection{Evaluation Configuration}
Every metric is computed over the fixed 1,000-image evaluation set. The results tables report cell means; the per-cell standard deviations are provided with the released raw outputs, and the per-image dispersion of the noisiest metric (FC) is analyzed explicitly in Section~\ref{sec:quant}. Statistical separation of the methods is tested with Bonferroni-corrected Friedman tests and Nemenyi post-hoc comparisons (Section~\ref{sec:stats}). To maintain a tractable computational grid, particularly for the highly expensive perturbation methods, we parameterized the \textsc{Quantus} evaluation metrics with fixed, standardized budgets:
\begin{itemize}
    \item \textbf{FC:} Configured for 20 perturbation runs per image, modifying feature subsets of size 224 at each step.
    \item \textbf{FE:} Configured to incrementally step and occlude 448 features per iteration.
    \item \textbf{MS:} Evaluated using 3 localized perturbation samples per image to measure spatial robustness.
\end{itemize}
The FC subset size of $224$ is the \textsc{Quantus} default; we reduce its run count to $20$ (from the default $100$) and use a single seed to keep the full $13\times8$ grid, including the expensive perturbation methods, tractable. The FC conclusion is therefore specific to this configuration, and a sweep over subset size and run count is left to future work (Section~\ref{sec:limitations}).

\FloatBarrier
\section{Results}
\label{sec:results}

\subsection{Quantitative Results and Architectural Failures}
\label{sec:quant}
To isolate the precise structural failure points of standard XAI tools, we pivot our evaluation to analyze specific metric axes across the evolutionary timeline of visual architectures. Table~\ref{tab:pg} and Table~\ref{tab:ms} present the Localization (PG) and Robustness (MS) scores, respectively. As vision backbones move away from flat patch-token processing to window-based or multi-scale linear attention, several standard attribution methods degrade. AttnLRP is reported only for the isotropic ViT-B/16, where its published softmax rules apply; on the other backbones, the toolbox has no attention-specific rule, so rather than report numbers from a generic fallback that would confound a transfer failure with a wrong-algorithm artifact, we mark those cells undefined (Section~\ref{sec:framework}).

\begin{table*}[t]
  \centering
  \caption{Localization: Pointing Game (PG $\uparrow$), scored against the object bounding box. Best per column in bold. Discussed in Section~\ref{sec:quant}.}
  \label{tab:pg}
  \setlength{\tabcolsep}{4pt}
  \begin{tabular}{l cc cc cc cc}
    \toprule
     & \multicolumn{1}{c}{\textbf{CNN}} & \multicolumn{1}{c}{\textbf{Isotropic}} & \multicolumn{2}{c}{\textbf{Hierarchical}} & \multicolumn{1}{c}{\textbf{Multi-axis}} & \multicolumn{1}{c}{\textbf{Hybrid}} & \multicolumn{2}{c}{\textbf{Linear Attn.}} \\
    \cmidrule(lr){2-2} \cmidrule(lr){3-3} \cmidrule(lr){4-5} \cmidrule(lr){6-6} \cmidrule(lr){7-7} \cmidrule(lr){8-9}
    \textbf{Method} & RN-50 & ViT-B/16 & Swin-B & PVT-v2 & MaxViT-S & MobileViT & EffViT-B1 & EffViT-B2 \\
    \midrule
    Saliency & 0.82 & 0.65 & 0.62 & 0.62 & 0.58 & 0.90 & 0.86 & 0.66 \\
    Integrated Grad & 0.92 & 0.64 & 0.66 & 0.64 & 0.62 & 0.90 & 0.90 & 0.90 \\
    Input$\times$Grad & 0.86 & 0.64 & 0.67 & 0.66 & 0.56 & 0.88 & 0.74 & 0.64 \\
    SmoothGrad & 0.97 & 0.85 & 0.79 & \best{0.92} & 0.70 & 0.92 & 0.92 & \best{0.92} \\
    VarGrad & \best{0.98} & 0.76 & 0.71 & 0.88 & 0.68 & 0.90 & 0.90 & \best{0.92} \\
    GradientSHAP & 0.92 & 0.70 & 0.73 & 0.70 & 0.65 & 0.88 & 0.86 & 0.84 \\
    \midrule
    Grad-CAM & \best{0.98} & \best{0.93} & \best{1.00} & 0.84 & \best{1.00} & \best{1.00} & 0.70 & 0.55 \\
    Grad-CAM++ & 0.97 & 0.49 & \best{1.00} & 0.88 & 0.80 & \best{1.00} & 0.44 & 0.56 \\
    \midrule
    Attention Rollout & \nd & 0.64 & \nd & \nd & \nd & \nd & \nd & \nd \\
    AttnLRP &\nd & 0.83 & \nd & \nd & \nd & \nd & \nd & \nd \\
    \midrule
    Occlusion & 0.95 & 0.76 & 0.78 & 0.78 & 0.66 & 0.92 & \best{0.96} & 0.88 \\
    RISE & 0.82 & 0.73 & 0.68 & 0.68 & 0.70 & 0.90 & 0.92 & 0.78 \\
    LIME & 0.94 & 0.87 & 0.86 & 0.86 & 0.80 & 0.90 & 0.80 & 0.90 \\
    \bottomrule
  \end{tabular}
\end{table*}

\begin{table*}[t]
  \centering
  \caption{\textbf{Dense-mask localization: Energy-Based Pointing Game (EBPG $\uparrow$).} Fraction of attribution energy inside the pixel-perfect ImageNet-S silhouette ($n=1000$); the random-map baseline is the mean mask area, $0.18$. Unlike the bounding-box Pointing Game (Table~\ref{tab:pg}), EBPG does not saturate: the ``perfect'' $1.00$ cells there fall to $\approx 0.5$ here, and no method exceeds $0.60$, so under a pixel-perfect target Grad-CAM's apparent dominance largely disappears and a gradient method (VarGrad) is competitive or better. Best per column in bold; computed for the gradient and CAM families.}
  \label{tab:ebpg}
  \setlength{\tabcolsep}{4pt}
  \begin{tabular}{l cc cc cc cc}
    \toprule
     & \multicolumn{1}{c}{\textbf{CNN}} & \multicolumn{1}{c}{\textbf{Isotropic}} & \multicolumn{2}{c}{\textbf{Hierarchical}} & \multicolumn{1}{c}{\textbf{Multi-axis}} & \multicolumn{1}{c}{\textbf{Hybrid}} & \multicolumn{2}{c}{\textbf{Linear Attn.}} \\
    \cmidrule(lr){2-2} \cmidrule(lr){3-3} \cmidrule(lr){4-5} \cmidrule(lr){6-6} \cmidrule(lr){7-7} \cmidrule(lr){8-9}
    \textbf{Method} & RN-50 & ViT-B/16 & Swin-B & PVT-v2 & MaxViT-S & MobileViT & EffViT-B1 & EffViT-B2 \\
    \midrule
    Saliency & 0.44 & 0.30 & 0.29 & 0.32 & 0.28 & 0.46 & 0.43 & 0.38 \\
    Input$\times$Grad & 0.44 & 0.31 & 0.31 & 0.35 & 0.29 & 0.46 & 0.44 & 0.39 \\
    SmoothGrad & 0.46 & 0.40 & 0.37 & 0.43 & 0.38 & 0.44 & 0.43 & 0.45 \\
    VarGrad & \best{0.58} & 0.40 & 0.39 & \best{0.53} & 0.44 & \best{0.55} & \best{0.54} & \best{0.57} \\
    Grad-CAM & \best{0.58} & \best{0.56} & \best{0.51} & 0.43 & \best{0.60} & \best{0.55} & 0.39 & 0.27 \\
    Grad-CAM++ & 0.51 & 0.44 & 0.50 & 0.42 & 0.43 & 0.51 & 0.28 & 0.30 \\
    \bottomrule
  \end{tabular}
\end{table*}

\subsubsection{Architectural evolution actively breaks localization}
\begin{figure}[!t]
  \centering
  \includegraphics[width=\linewidth]{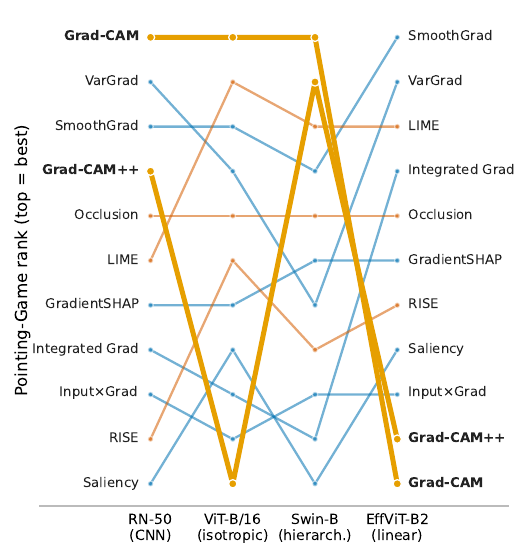}
  \caption{\textbf{Rankings do not transfer.} Pointing Game rank of the ten methods applicable to every backbone, across the architectural evolution from CNN to linear attention. Grad-CAM (bold orange) holds rank 1 on the CNN, isotropic, and hierarchical backbones, then falls to last place on the linear-attention EfficientViT-B2; Grad-CAM++ swings between rank 2 and rank 10 depending on the backbone. Ranks use the bounding-box Pointing Game; under the dense-mask EBPG (Table~\ref{tab:ebpg}), the ordering shifts further, lifting VarGrad and lowering Grad-CAM.}
  \label{fig:rankflip}
\end{figure}

On ResNet-50, the PG is effectively saturated, with eight of the eleven applicable methods scoring at least $0.92$. This saturation is intrinsic to the metric rather than a property of any method. Because the target is the bounding box of the object and that box covers a large, typically central image region, any compact and center-weighted map places its single peak inside the box on nearly every image, so a handful of cells reach the ceiling of $1.00$ (Grad-CAM on Swin-B, MaxViT-S, and MobileViT-v2; Grad-CAM++ on Swin-B and MobileViT-v2). A score of $1.00$ thus certifies only that the peak never left a large central rectangle, not that the map traces the object silhouette. Crucially, this is not an indexing artifact that artificially pins the metric near the ceiling: the identical code path returns well-separated and sub-chance scores elsewhere in the same table, for instance Grad-CAM $0.55$ and Grad-CAM++ $0.44$ on EfficientViT, both below the $0.61$ random-point floor, which a metric locked near $1.00$ could never produce. The same routine emits the ceiling and the floor, and the object-mask bounding box, not the whole-image frame, is what defines a hit. On the transformer backbones, the performance landscape fractures (Figure~\ref{fig:pointing}). Grad-CAM remains a strong localizer across the CNN, isotropic ($0.93$ on ViT-B/16), hierarchical (Swin-B $1.00$, PVT-v2 $0.84$), multi-axis (MaxViT-S $1.00$), and even the convolution-hybrid MobileViT-v2 ($1.00$) backbones, but collapses specifically on the linear-attention EfficientViT variants, dropping to $0.70$ on B1 and $0.55$ on B2. EfficientViT still exposes a $7 \times 7$ terminal spatial feature map (we verified the tensor shape, identical in resolution to Swin and PVT-v2), so the failure is not a missing feature map. Its multi-scale linear attention instead gives every spatial location a global receptive field: we measure the terminal feature energy to be markedly flatter than Swin's (spatial coefficient of variation $0.29$ versus $0.56$ over $20$ images), so the gradient-weighted CAM forms a sharp but mislocalized peak that no longer lands on the object. Grad-CAM++ is even less reliable: it scores $0.97$ on the CNN yet collapses to $0.49$ on the isotropic ViT-B/16 and to $0.44$ and $0.56$ on EfficientViT-B1 and -B2, and even on the multi-axis MaxViT-S its global-grid instability (Section~\ref{sec:gradcam_collapse}) leaves it at $0.80$ where plain Grad-CAM reaches $1.00$. Meanwhile, plain gradient methods (Saliency, IG) that score well on CNNs fall on the isotropic, hierarchical, and multi-axis transformers (into the $0.58$--$0.67$ range), though they recover on the convolution-hybrid MobileViT-v2. These rank-flips, traced method by method in Figure~\ref{fig:rankflip}, indicate that architectural design choices, and linear attention in particular, strongly shape XAI performance. To quantify the (non-)transfer against clear baselines, we compute the mean pairwise Spearman correlation between the method rankings of the eight backbones. Perfect transfer would give $1.0$; a permutation null with independently shuffled rankings gives $0.00 \pm 0.06$. We observe $0.35$ (95\% bootstrap CI $[0.21, 0.48]$), significantly above the random null ($p < 10^{-4}$) yet far below perfect transfer, and it falls to $0.17$ between the CNN and the linear-attention EfficientViT-B2. Rankings therefore transfer only weakly and partially, not cleanly, and are least related across the architectural extremes.

To remove the bounding-box confound entirely, Table~\ref{tab:ebpg} rescores localization with the Energy-Based Pointing Game (EBPG), the fraction of attribution energy that falls inside the pixel-perfect ImageNet-S silhouette, against a random-map floor of $0.18$ (the mean mask area). Three things change. First, the saturation vanishes: the bounding-box $1.00$ cells fall to $0.51$ (Swin-B), $0.60$ (MaxViT-S), and $0.55$ (MobileViT-v2), confirming that those maps place only about half their energy on the object. Second, no method is a strong dense localizer: EBPG never exceeds $0.60$, so every method leaves a large share of energy off-object. Third, Grad-CAM's apparent dominance under the bounding box is largely an artifact of that metric: under EBPG, the gradient-based VarGrad ties or beats Grad-CAM on five of the eight backbones, while Grad-CAM's linear-attention collapse is sharper still (EfficientViT-B2 $0.27$, barely above the floor). The bounding-box and dense metrics therefore agree about \emph{failure} (EfficientViT), but the bounding box badly overstates \emph{success}; we treat EBPG as the localization metric of record and retain the bounding-box Pointing Game only for comparison with prior work.

\begin{table*}[t]
  \centering
  \caption{Robustness: Max-Sensitivity (MS $\downarrow$). A lower value indicates greater stability; the best score in each column is highlighted in bold. The values represent averages from a single seed; some cells are influenced by a heavy right tail (notably VarGrad on Swin-B and Grad-CAM++ on ViT-B/16). For these cases, the outlier-robust medians presented in Table~\ref{tab:cam_ms} offer more dependable summaries. Further discussion is provided in Section~\ref{sec:quant}.}
  \label{tab:ms}
  \setlength{\tabcolsep}{4pt}
  \begin{tabular}{l cc cc cc cc}
    \toprule
     & \multicolumn{1}{c}{\textbf{CNN}} & \multicolumn{1}{c}{\textbf{Isotropic}} & \multicolumn{2}{c}{\textbf{Hierarchical}} & \multicolumn{1}{c}{\textbf{Multi-axis}} & \multicolumn{1}{c}{\textbf{Hybrid}} & \multicolumn{2}{c}{\textbf{Linear Attn.}} \\
    \cmidrule(lr){2-2} \cmidrule(lr){3-3} \cmidrule(lr){4-5} \cmidrule(lr){6-6} \cmidrule(lr){7-7} \cmidrule(lr){8-9}
    \textbf{Method} & RN-50 & ViT-B/16 & Swin-B & PVT-v2 & MaxViT-S & MobileViT & EffViT-B1 & EffViT-B2 \\
    \midrule
    Saliency & 0.770 & 2.692 & 1.240 & 0.892 & 1.123 & 0.804 & 0.787 & 0.930 \\
    Integrated Grad & 0.887 & 1.110 & 1.114 & 1.085 & 1.211 & 0.858 & 0.913 & 1.745 \\
    Input$\times$Grad & 0.905 & 3.291 & 1.390 & 1.035 & 1.324 & 0.917 & 0.955 & 1.061 \\
    SmoothGrad & 0.242 & 0.699 & 0.638 & 0.254 & 0.500 & 0.275 & 0.270 & 0.299 \\
    VarGrad & 0.607 & 16.413 & 73.493 & 0.858 & 3.070 & 0.805 & 0.848 & 0.743 \\
    GradientSHAP & 0.963 & 1.420 & 1.610 & 1.250 & 1.400 & 0.950 & 1.050 & 1.120 \\
    \midrule
    Grad-CAM & 0.273 & 1.939 & \best{0.204} & 0.182 & \best{0.289} & \best{0.211} & 1.194 & 1.700 \\
    Grad-CAM++ & \best{0.189} & 17.176 & 0.206 & \best{0.174} & 1.740 & 0.226 & 1.174 & 1.677 \\
    \midrule
    Attention Rollout & \nd & \best{0.090} & \nd & \nd & \nd & \nd & \nd & \nd \\
    AttnLRP &\nd & 0.926 & \nd & \nd & \nd & \nd & \nd & \nd \\
    \midrule
    Occlusion & 0.938 & 0.995 & 1.449 & 0.883 & 1.073 & 1.019 & 0.709 & 0.953 \\
    RISE & 0.327 & 0.186 & 0.273 & 0.280 & 0.325 & 0.444 & \best{0.266} & \best{0.220} \\
    LIME & 1.312 & 1.677 & 1.896 & 1.787 & 3.179 & 1.501 & 1.266 & 1.629 \\
    \bottomrule
  \end{tabular}
\end{table*}

\begin{figure}[!t]
  \centering
  \includegraphics[width=1\columnwidth]{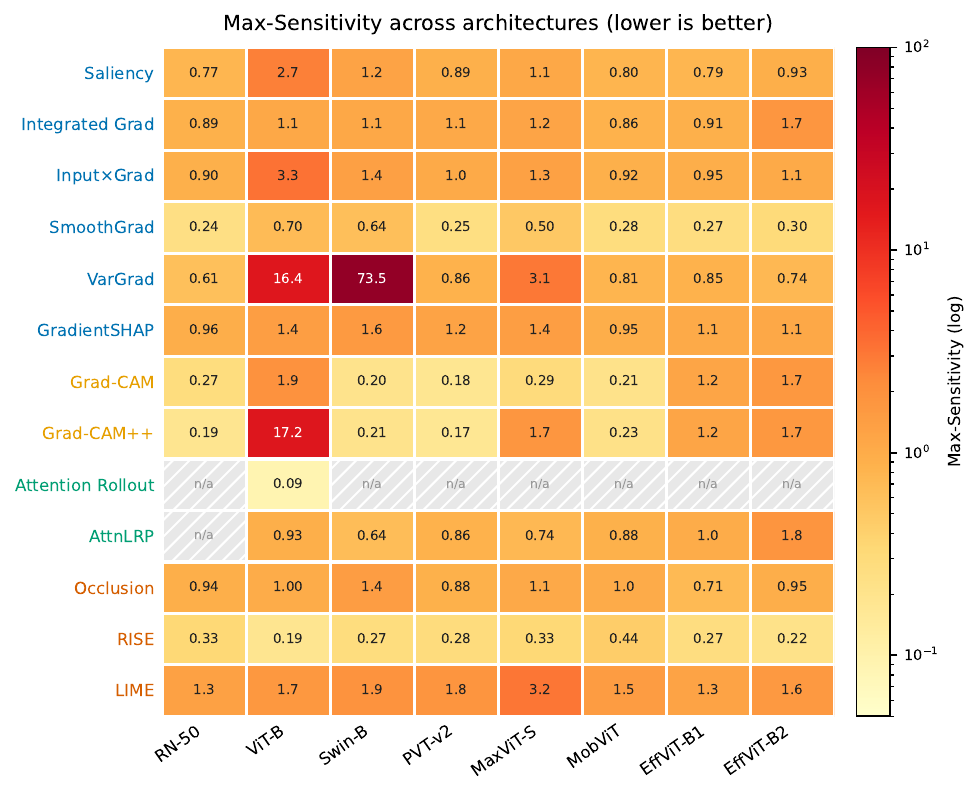}
  \caption{\textbf{Robustness degrades unevenly on transformers.} MS (log color; darker $=$ less stable). Some methods are stable on the CNN spike by orders of magnitude on specific backbones, notably VarGrad on Swin-B ($73.5$) and Grad-CAM++ on ViT-B/16 ($17.2$).}
  \label{fig:robust}
\end{figure}

\subsubsection{Robustness varies widely across model architectures}
MS exposes large instabilities in some standard methods on transformers, though the effect is backbone-specific rather than uniform. Attention Rollout is highly stable ($0.090$ on ViT-B/16), but at the cost of nearly flat, unlocalized maps. Grad-CAM++ is severely unstable on the global-attention ViT-B/16, where its MS reaches $17.18$ against a well-behaved $0.19$ on the CNN, yet it is as stable as Grad-CAM on windowed Swin-B (Section~\ref{sec:gradcam_collapse}). VarGrad, whose estimator is the variance of perturbed gradients and is therefore heavy-tailed by construction, is instead highly volatile on Swin-B ($73.5$, an $85\times$ jump from PVT-v2's $0.86$); as with Grad-CAM++, a handful of images with extreme gradient variance dominate this single-seed mean, and multi-seed intervals would tighten it. A plausible mechanism for the Swin-specific spike, which we offer as a hypothesis rather than a measured cause, is its shifted-window partitioning: under the input noise that VarGrad injects, tokens near a window edge can cross the hard window boundary, producing large discontinuous jumps in the per-token gradient that the variance estimator amplifies, whereas PVT-v2's smooth spatial-reduction pooling introduces no such boundaries. Which method destabilizes depends on the backbone, and gradient-based and CAM-based techniques are the most affected (Figure~\ref{fig:robust}).

\begin{table*}[t]
  \centering
  \caption{Faithfulness: Faithfulness Correlation (FC $\uparrow$). Values cluster near zero across the grid; analyzed in Section~\ref{sec:quant}.}
  \label{tab:fc}
  \setlength{\tabcolsep}{4pt}
  \begin{tabular}{l cc cc cc cc}
    \toprule
     & \multicolumn{1}{c}{\textbf{CNN}} & \multicolumn{1}{c}{\textbf{Isotropic}} & \multicolumn{2}{c}{\textbf{Hierarchical}} & \multicolumn{1}{c}{\textbf{Multi-axis}} & \multicolumn{1}{c}{\textbf{Hybrid}} & \multicolumn{2}{c}{\textbf{Linear Attn.}} \\
    \cmidrule(lr){2-2} \cmidrule(lr){3-3} \cmidrule(lr){4-5} \cmidrule(lr){6-6} \cmidrule(lr){7-7} \cmidrule(lr){8-9}
    \textbf{Method} & RN-50 & ViT-B/16 & Swin-B & PVT-v2 & MaxViT-S & MobileViT & EffViT-B1 & EffViT-B2 \\
    \midrule
    Saliency & -0.006 & -0.027 & 0.001 & 0.048 & -0.008 & 0.018 & -0.027 & 0.018 \\
    Integrated Grad & 0.005 & 0.012 & \best{0.046} & \best{0.078} & 0.041 & -0.011 & -0.047 & 0.009 \\
    Input$\times$Grad & 0.012 & -0.019 & 0.045 & 0.016 & 0.024 & -0.037 & -0.018 & \best{0.058} \\
    SmoothGrad & -0.005 & -0.004 & 0.042 & -0.014 & -0.072 & 0.023 & -0.030 & 0.007 \\
    VarGrad & -0.024 & -0.026 & 0.020 & 0.042 & 0.021 & -0.030 & -0.041 & -0.046 \\
   GradientSHAP & -0.014 & 0.018 & 0.005 & 0.050 & 0.030 & -0.010 & -0.020 & 0.015 \\
    \midrule
    Grad-CAM & 0.016 & -0.001 & 0.033 & -0.045 & -0.054 & -0.001 & 0.042 & 0.041 \\
    Grad-CAM++ & \best{0.054} & \best{0.023} & -0.018 & -0.041 & \best{0.068} & -0.036 & \best{0.085} & -0.007 \\
    \midrule
    Attention Rollout & \nd & -0.023 & \nd & \nd & \nd & \nd & \nd & \nd \\
    AttnLRP &\nd & -0.009 & \nd & \nd & \nd & \nd & \nd & \nd \\
    \midrule
    Occlusion & -0.007 & 0.001 & -0.024 & 0.015 & 0.047 & \best{0.030} & 0.050 & 0.006 \\
    RISE & -0.003 & -0.006 & -0.041 & 0.020 & 0.002 & -0.002 & -0.013 & -0.015 \\
    LIME & 0.013 & -0.014 & -0.001 & -0.005 & 0.014 & 0.012 & -0.011 & -0.038 \\
    \bottomrule
  \end{tabular}
\end{table*}

\begin{table*}[t]
  \centering
  \caption{Faithfulness: Faithfulness Estimate (FE $\uparrow$). More discriminative than FC but sign-flips across backbones; analyzed in Section~\ref{sec:quant}.}
  \label{tab:fe}
  \setlength{\tabcolsep}{4pt}
  \begin{tabular}{l cc cc cc cc}
    \toprule
     & \multicolumn{1}{c}{\textbf{CNN}} & \multicolumn{1}{c}{\textbf{Isotropic}} & \multicolumn{2}{c}{\textbf{Hierarchical}} & \multicolumn{1}{c}{\textbf{Multi-axis}} & \multicolumn{1}{c}{\textbf{Hybrid}} & \multicolumn{2}{c}{\textbf{Linear Attn.}} \\
    \cmidrule(lr){2-2} \cmidrule(lr){3-3} \cmidrule(lr){4-5} \cmidrule(lr){6-6} \cmidrule(lr){7-7} \cmidrule(lr){8-9}
    \textbf{Method} & RN-50 & ViT-B/16 & Swin-B & PVT-v2 & MaxViT-S & MobileViT & EffViT-B1 & EffViT-B2 \\
    \midrule
    Saliency & 0.138 & -0.050 & \best{0.159} & \best{0.228} & \best{0.307} & 0.044 & 0.114 & \best{0.156} \\
    Integrated Grad & 0.035 & 0.016 & 0.079 & 0.213 & -0.056 & 0.011 & 0.076 & 0.028 \\
    Input$\times$Grad & -0.001 & -0.083 & 0.053 & 0.138 & 0.086 & 0.043 & 0.002 & 0.073 \\
    SmoothGrad & 0.107 & -0.034 & 0.030 & -0.269 & -0.328 & -0.037 & 0.132 & 0.085 \\
    VarGrad & 0.082 & -0.039 & -0.040 & -0.249 & -0.208 & 0.021 & 0.099 & 0.031 \\
   GradientSHAP & 0.034 & 0.025 & 0.070 & 0.210 & 0.050 & 0.020 & 0.080 & 0.060 \\
    \midrule
    Grad-CAM & \best{0.229} & 0.013 & 0.052 & -0.005 & -0.093 & -0.005 & 0.149 & 0.114 \\
    Grad-CAM++ & 0.178 & 0.015 & 0.051 & -0.018 & 0.063 & 0.005 & 0.062 & 0.047 \\
    \midrule
    Attention Rollout & \nd & -0.120 & \nd & \nd & \nd & \nd & \nd & \nd \\
    AttnLRP &\nd & -0.025 & \nd & \nd & \nd & \nd & \nd & \nd \\
    \midrule
    Occlusion & 0.160 & 0.079 & -0.045 & 0.008 & 0.153 & -0.014 & 0.234 & 0.057 \\
    RISE & 0.170 & \best{0.089} & 0.027 & -0.009 & 0.065 & \best{0.134} & \best{0.243} & 0.151 \\
    LIME & 0.015 & -0.051 & -0.248 & -0.255 & -0.162 & 0.037 & 0.097 & 0.038 \\
    \bottomrule
  \end{tabular}
\end{table*}

\subsubsection{Proxy FC does not discriminate under the standard protocol}
\begin{figure}[!t]
  \centering
  \includegraphics[width=\linewidth]{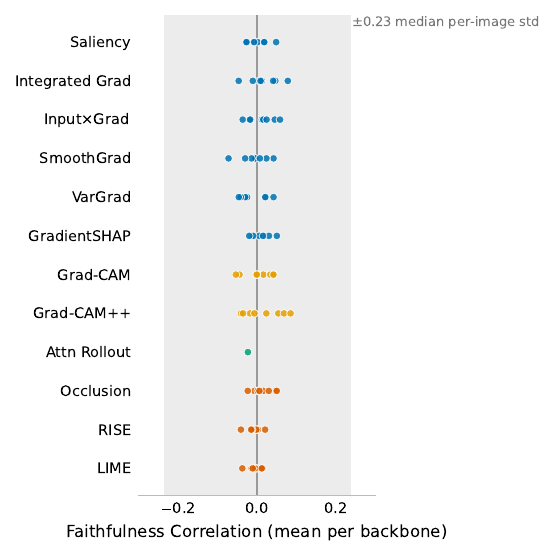}
  \caption{\textbf{FC behaves as statistical noise.} Per-backbone FC means per method (dots, by family) against the median per-image standard deviation ($\pm 0.234$). Every mean lies inside the noise band, so the metric cannot rank methods.}
  \label{fig:fcnoise}
\end{figure}

FC values cluster tightly around zero across the entire grid (Table~\ref{tab:fc} and Figure~\ref{fig:fcnoise}); the full range spans merely $-0.072$ to $+0.085$. The per-image standard deviations (roughly $0.22$ to $0.25$) are an order of magnitude larger than the between-method differences, so a single-image FC score is individually unreliable. The aggregate test agrees and is the proper basis for the claim: the Friedman omnibus for FC cannot reject identical method performance ($p = 0.57$, Section~\ref{sec:stats}), so FC does not rank methods here. We report this for the standard configuration (subset size $224$, $20$ runs); a configuration sweep could shift the magnitude, but the near-zero, non-separating behavior is what a practitioner obtains with the default protocol. The FE (Table~\ref{tab:fe}) offers slightly more discriminative power (e.g., Saliency scoring $0.307$ on MaxViT-S), but it exhibits contradictory behavior across architectures, changing sign for Grad-CAM between ResNet ($+0.229$) and MaxViT-S ($-0.093$). Proxy faithfulness correlation is therefore unreliable for cross-architecture evaluation under this standard protocol.

\begin{table*}[t]
  \centering
  \caption{Complexity: Sparseness (SP $\uparrow$). Higher is more concentrated; best per column in bold. Discussed in Section~\ref{sec:quant}.}
  \label{tab:sp}
  \setlength{\tabcolsep}{4pt}
  \begin{tabular}{l cc cc cc cc}
    \toprule
     & \multicolumn{1}{c}{\textbf{CNN}} & \multicolumn{1}{c}{\textbf{Isotropic}} & \multicolumn{2}{c}{\textbf{Hierarchical}} & \multicolumn{1}{c}{\textbf{Multi-axis}} & \multicolumn{1}{c}{\textbf{Hybrid}} & \multicolumn{2}{c}{\textbf{Linear Attn.}} \\
    \cmidrule(lr){2-2} \cmidrule(lr){3-3} \cmidrule(lr){4-5} \cmidrule(lr){6-6} \cmidrule(lr){7-7} \cmidrule(lr){8-9}
    \textbf{Method} & RN-50 & ViT-B/16 & Swin-B & PVT-v2 & MaxViT-S & MobileViT & EffViT-B1 & EffViT-B2 \\
    \midrule
    Saliency & 0.461 & 0.640 & 0.593 & 0.542 & 0.515 & 0.474 & 0.445 & 0.435 \\
    Integrated Grad & 0.616 & 0.650 & 0.654 & 0.612 & 0.608 & 0.614 & 0.612 & 0.606 \\
    Input$\times$Grad & 0.626 & 0.677 & 0.711 & 0.661 & 0.680 & 0.643 & 0.619 & 0.624 \\
    SmoothGrad & 0.344 & 0.406 & 0.297 & 0.312 & 0.270 & 0.362 & 0.311 & 0.335 \\
    VarGrad & 0.593 & 0.783 & 0.729 & 0.579 & 0.590 & 0.647 & 0.579 & 0.614 \\
    GradientSHAP & 0.616 & 0.660 & 0.670 & 0.620 & 0.650 & 0.630 & 0.640 & 0.650 \\
    \midrule
    Grad-CAM & 0.597 & 0.706 & 0.461 & 0.338 & 0.633 & 0.496 & 0.745 & 0.752 \\
    Grad-CAM++ & 0.471 & 0.593 & 0.453 & 0.310 & 0.697 & 0.452 & \best{0.784} & 0.788 \\
    \midrule
    Attention Rollout & \nd & 0.126 & \nd & \nd & \nd & \nd & \nd & \nd \\
    AttnLRP &\nd & 0.692 & \nd & \nd & \nd & \nd & \nd & \nd \\
    \midrule
    Occlusion & 0.509 & 0.476 & 0.366 & 0.430 & 0.392 & 0.498 & 0.481 & 0.479 \\
    RISE & 0.034 & 0.023 & 0.022 & 0.023 & 0.024 & 0.036 & 0.033 & 0.030 \\
    LIME & \best{0.760} & \best{0.870} & \best{0.864} & \best{0.818} & \best{0.895} & \best{0.781} & 0.775 & \best{0.811} \\
    \bottomrule
  \end{tabular}
\end{table*}

\subsubsection{Statistical significance of the method separation}
\label{sec:stats}
Following the protocol of Dem\v{s}ar~\cite{demsar2006statistical}, we treat the eight backbones as blocks, rank the eleven methods with complete backbone coverage within each block, and run one Friedman test per metric with a Bonferroni correction across the five metrics. Attention Rollout and AttnLRP are excluded from the test because the Friedman procedure requires complete blocks, and both methods are undefined on the CNN. The results are reported in Table~\ref{tab:friedman}. Localization (PG), robustness (MS), and complexity (SP) separate the methods decisively, with corrected $p$-values below $10^{-3}$. FC fails the omnibus test outright ($p = 0.57$ before any correction): the data cannot reject the hypothesis that all methods perform identically under FC, which elevates the noise finding above from an observation to a formal statistical statement. FE is intermediate; its uncorrected $p = 0.011$ loses significance after correction ($p_{\text{Bonf}} = 0.055$). Post-hoc Nemenyi comparisons at $\alpha = 0.05$ give a critical difference of $\mathrm{CD} = 5.34$ mean-rank units ($k = 11$ methods, $N = 8$ blocks); for each significant metric, the table lists the methods whose mean rank falls beyond the critical difference from the best-ranked method.

With only $N = 8$ backbones serving as blocks, this design is deliberately conservative. The critical difference of $5.34$ spans nearly half of the eleven-method rank scale, so the post-hoc test can certify only coarse separations between the top-ranked method and the tail, not fine-grained differences between methods at adjacent ranks. We therefore attach significance only to the large effects the test actually resolves: the SmoothGrad advantage over Saliency and Input$\times$Gradient in localization, the RISE advantage over GradientSHAP and LIME in robustness, the LIME sparseness lead, and the FC omnibus null. Small rank differences elsewhere in Tables~\ref{tab:pg}--\ref{tab:sp} are reported descriptively, not as statistically separated methods, and the same caution applies to the practical recommendations in Section~\ref{sec:findings}, which rest on large score gaps and rank swings rather than adjacent-rank ordering. Tightening the finer comparisons requires more blocks, that is, more backbones within each family, which is the primary axis along which we are extending the benchmark. Two further caveats cut in the conservative direction. The near-miss on FE ($p_{\mathrm{Bonf}} = 0.055$) should be read as insufficient power, a likely Type II outcome, not as evidence that methods perform identically under FE; only FC, with its far larger uncorrected $p = 0.57$, supports a genuine null. And the Bonferroni correction assumes independent metrics, whereas faithfulness, localization, and robustness measure related properties of the same maps, so the corrected $p$-values are conservative. We therefore treat FE as inconclusive rather than null.

\begin{table}[!t]
\centering
\caption{Friedman test results across the $11\times8$ method--backbone grid. Bonferroni correction was applied across the five metrics. Post-hoc Nemenyi comparisons are reported only when the corrected omnibus test is significant ($\mathrm{CD}=5.34$, $\alpha=0.05$).}
\label{tab:friedman}
\footnotesize
\setlength{\tabcolsep}{3pt}
\renewcommand{\arraystretch}{1.05}

\begin{tabular}{lcccll}
\toprule
Metric & $\chi^2_F$ & $p$ & $p_{\mathrm{Bonf}}$ & Best & Beyond CD \\
\midrule
PG & 34.6 & $1.5\times10^{-4}$ & $7.4\times10^{-4}$ & SG (2.94) & Sal., I$\times$G \\
MS & 36.8 & $6.3\times10^{-5}$ & $3.1\times10^{-4}$ & RISE (2.50) & GS, LIME \\
FC & 8.6 & 0.571 & 1.000 & -- & -- \\
FE & 22.9 & 0.011 & 0.055 & -- & -- \\
SP & 61.6 & $1.8\times10^{-9}$ & $9.1\times10^{-9}$ & LIME (1.12) & Sal., SG, Occ., RISE \\
\bottomrule
\end{tabular}

\vspace{2pt}
\raggedright
\footnotesize
\textit{Abbreviations:} Sal.=Saliency, I$\times$G=Input$\times$Gradient, GS=GradientSHAP, SG=SmoothGrad, Occ.=Occlusion.
\end{table}

\subsubsection{Complexity trade-off}

\begin{table}[!t]
  \centering
  \caption{Mean computational cost per explanation (ms), averaged over backbones
  and the 1,000-image set. $^{\dagger}$Attention Rollout and AttnLRP average over ViT-B/16
  only, the sole backbone where they apply, so their times are indicative and not strictly
  comparable to methods averaged over all eight backbones. Perturbation methods
  dominate the budget.}
  \label{tab:cost}
  \setlength{\tabcolsep}{6pt}
  \begin{tabular}{@{}llr@{}}
    \toprule
    Method & Family & Time (ms) \\
    \midrule
    SmoothGrad            & Gradient      & 20.9 \\
    Input$\times$Gradient & Gradient      & 22.1 \\
    Grad-CAM++            & CAM           & 24.0 \\
    Grad-CAM              & CAM           & 26.5 \\
    Saliency              & Gradient      & 31.9 \\
    GradientSHAP          & Gradient      & 32.7 \\
    VarGrad               & Gradient      & 40.4 \\
    Attention Rollout$^{\dagger}$     & Attention     & 41.0 \\
    AttnLRP$^{\dagger}$               & Attention     & 57.4 \\
    IG  & Gradient      & 479.6 \\
    RISE                  & Perturbation  & 669.9 \\
    LIME                  & Perturbation  & 1\,488.1 \\
    Occlusion             & Perturbation  & 11\,011.1 \\
    \bottomrule
  \end{tabular}
\end{table}

SP orders the methods by concentration. LIME produces the sparsest attributions ($0.76$--$0.90$), while RISE produces highly diffuse, scattered maps (failing to exceed $0.13$). Importantly, SP conflates concentration with native resolution: LIME and Grad-CAM operate on coarse super-pixels or a $7 \times 7$ grid upsampled to $224 \times 224$, whereas Saliency and IG produce dense pixel-level maps, so a low-resolution method can score as ``sparser'' simply because its support is coarser rather than because it is more selective. SP should therefore be read as a descriptive trait, not a quality ranking, and comparisons across methods of different native resolution are confounded. The decisive Friedman separation for SP (Table~\ref{tab:friedman}) reflects this native-resolution difference as much as any quality difference, so we do not treat it as a cross-method quality claim.

\subsubsection{Cost and fidelity trade-off}
\begin{figure*}[t]
  \centering
  \includegraphics[width=0.72\textwidth]{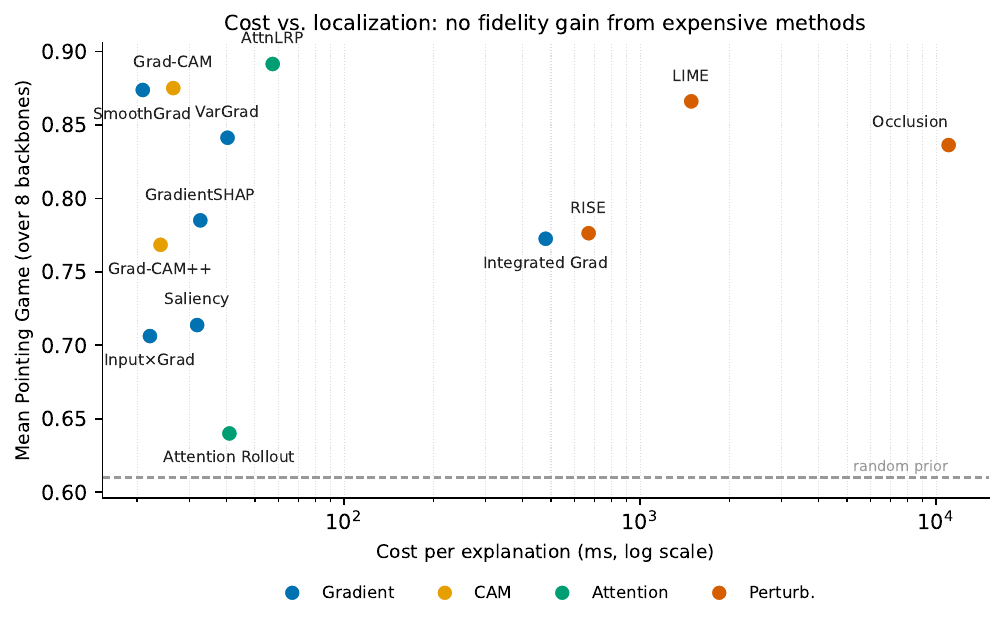}
  \caption{\textbf{Cost buys no fidelity.} Mean PG (over eight backbones) versus wall-clock cost per explanation (log scale), by family. Cheap CAM and gradient methods match or beat perturbation methods two to three orders of magnitude slower.}
  \label{fig:pareto}
\end{figure*}

Table~\ref{tab:cost} reports the mean wall-clock cost per explanation. Gradient-based and CAM-based methods are highly efficient ($21$--$40$\, ms), attention-native methods are moderate ($41$--$57$\, ms), while perturbation methods completely dominate the computational budget. IG takes $0.48$\,s, RISE takes $0.67$\,s, and Occlusion takes an unmanageable $11.0$\,s per image, a $\sim$500$\times$ premium over SmoothGrad. Because Grad-CAM is among the cheapest yet provides strong localization on the CNN and softmax-attention ViTs, the evaluation suite exposes a steep cost-vs-fidelity frontier where expensive perturbation methods yield little practical gain (Figure~\ref{fig:pareto}).

\subsection{Qualitative Results}
\label{sec:qual}

\begin{figure*}[!t]
  \centering
  \includegraphics[width=\textwidth]{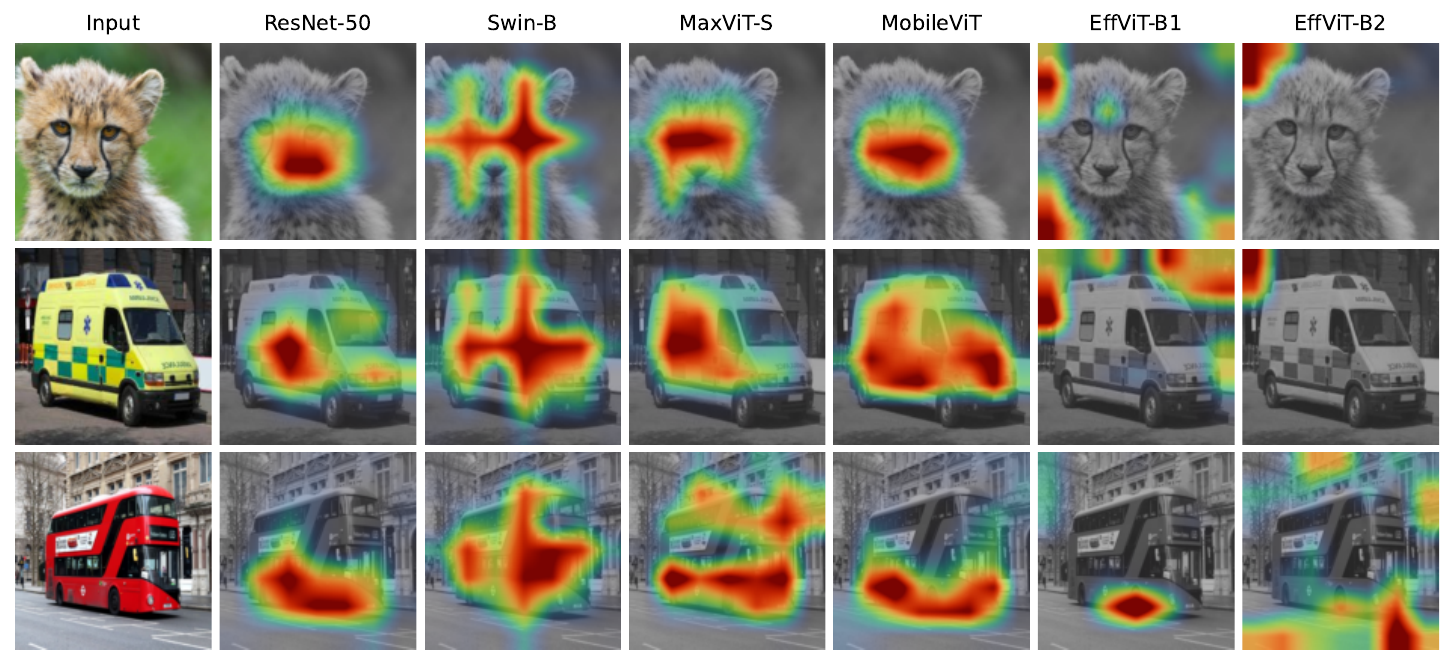}
  \caption{\textbf{Grad-CAM across architectures.} Heatmaps on three ImageNet-S images stay on the object for the CNN, hierarchical, multi-axis, and hybrid backbones, then scatter to the background on the linear-attention EfficientViT variants, the visual analog of the Pointing Game collapse in Table~\ref{tab:pg}.}
  \label{fig:qualitative}
\end{figure*}

The qualitative picture mirrors the quantitative one (Figure~\ref{fig:qualitative}).
Grad-CAM yields compact, object-centered maps on the CNN, hierarchical, multi-axis,
and convolution-hybrid backbones, but on the linear-attention EfficientViT variants
its relevance scatters to the image corners and background, missing the object
entirely; on Swin-B and PVT-v2 the coarse $7 \times 7$ terminal grid shows through
as a block artifact (Section~\ref{sec:swin_bottlenecks}). This visually confirms the
Pointing Game collapse of Table~\ref{tab:pg}: EfficientViT retains a spatial feature
map, but its linear attention diffuses the terminal features so the gradient-weighted
CAM mislocalizes (Section~\ref{sec:quant}).
Attention rollout, by contrast, is diffuse and spreads probability mass across
background tokens (consistent with its low sparseness and weak PG score),
while perturbation methods such as LIME produce sharp but blocky super-pixel
attributions.

Figure~\ref{fig:qual-grid-cat} condenses the full failure landscape into a single
view: one input, all eight backbones, thirteen methods. The failure instances are
architectural rather than random. The Grad-CAM family collapses exactly on the
linear-attention rows and rollout collapses exactly on the global-attention row,
while Captum LRP (not shown) runs on no backbone at all, rejecting a different
unsupported module on each family. No method family survives every architecture.

\FloatBarrier

\section{Discussion}
\label{sec:discussion}

\subsection{Key Findings}
\label{sec:findings}
\textbf{(1) XAI rankings do not transfer across architectures.} The implicit assumption that attribution methods scoring highly on CNNs remain optimal for ViTs is empirically false. Rankings do not merely rearrange; large failures emerge. Grad-CAM++ falls from a top localizer on the CNN (PG $0.97$) to below the $0.61$ random-point floor on the isotropic ViT-B/16 (PG $0.49$), and baseline Grad-CAM drops from $0.98$ on the CNN to $0.55$ on the linear-attention EfficientViT, also below that floor. Evaluating an XAI tool solely on ResNet is no longer a valid proxy for modern vision models.

\textbf{(2) The standard faithfulness-correlation protocol does not discriminate methods at this scale.} Under the community-standard FC configuration, the metric does not separate strong from weak explanations: it is the only metric whose Friedman omnibus test cannot reject the hypothesis that all methods perform identically, even before correction ($p = 0.57$, Section~\ref{sec:stats}). We stress that this is a statement about the standard protocol as applied, not a universal impossibility claim for faithfulness. It argues for pivoting away from single-metric occlusion correlation and toward ground-truth localization, causal benchmarks, or Shapley-based approximations.

\textbf{(3) Architectural evolution destabilizes XAI unevenly.} As ViTs evolve from isotropic patch tokens to hierarchical and multi-scale designs, gradient flow and structural properties change substantially. MS exposes this, but the instability is backbone-specific rather than uniform. The higher-order Grad-CAM++ explodes on the global-softmax ViT-B/16 (MS $17.18$) yet is stable and nearly indistinguishable from Grad-CAM on windowed Swin-B, while the variance-based VarGrad spikes to $73.5$ on Swin-B, up to two orders of magnitude above its CNN value. Which method breaks depends on which architectural bottleneck its estimator is most exposed to (Section~\ref{sec:gradcam_collapse}).

\begin{figure*}[!t]
  \centering
  \includegraphics[width=\textwidth]{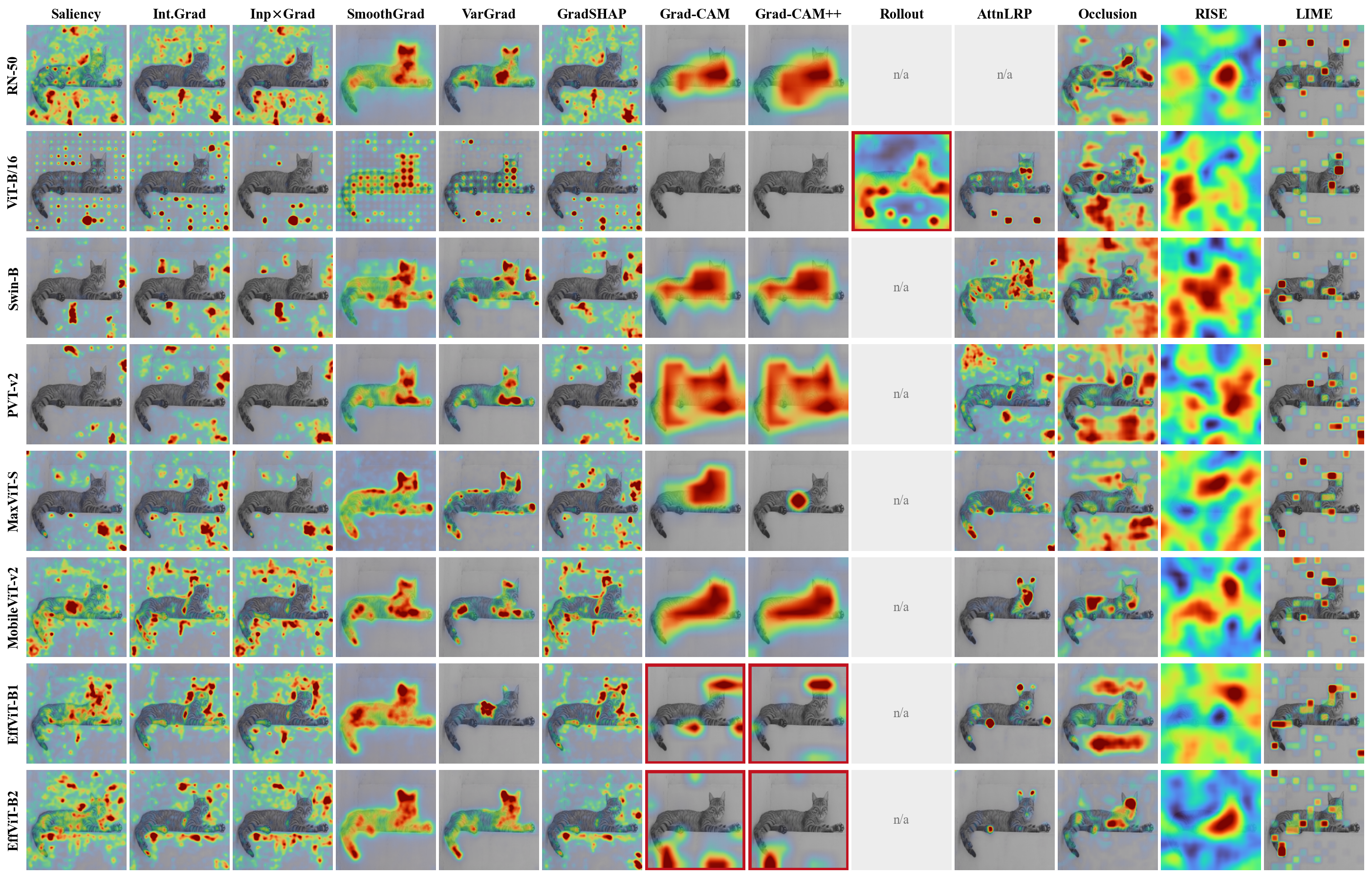}
  \caption{\textbf{Full method-by-backbone gallery} (one input, all backbones and methods, shared percentile normalization). Red-bordered cells mark documented failures: the Grad-CAM family loses the object on the linear-attention EfficientViT variants, and attention rollout is near content-free on ViT-B, consistent with Table~\ref{tab:pg}. Gray cells are structurally undefined.}
  \label{fig:qual-grid-cat}
\end{figure*}

\textbf{(4) Cost and fidelity are in tension.} Perturbation methods like LIME and Occlusion are prohibitively expensive (up to $\sim$500$\times$ slower than gradients) but do not buy a proportionate increase in localization or stability. In practical terms, Grad-CAM remains a computationally cheap and generally reliable default for ViTs, except on models employing linear attention, where its spatial-map assumptions collapse.

\subsection{Theoretical Analysis: Why Grad-CAM++ Collapses under Global Attention}
\label{sec:gradcam_collapse}
Grad-CAM++ assumes that the pre-softmax class score $Y^c$ (the logit of the target class $c$, identical to $F_c(x)$ in Section~\ref{sec:attribution_methods}) is a linear combination of spatial feature maps $A^k$ via GAP, such that $Y^c = \sum_k w^c_k \sum_{i,j} A^k_{i,j}$, where $w^c_k$ is the class-specific weight of feature map $k$. Under this assumption, the second- and third-order derivatives used to compute the pixel-wise weighting coefficients $\alpha_{i,j}^{kc}$ (the weight Grad-CAM++ assigns to the positive partial derivative at location $(i,j)$ of map $k$ when forming $w^c_k$) are stable and positive.

In a standard ViT, $Y^c$ is not a global average pool of a terminal feature map. It is read from a classification token $z_{\text{cls}}$ that attends to the spatial tokens through a nonlinear softmax:
\begin{equation}
z_{\text{cls}} = \sum_i \text{softmax}\!\left(\frac{q_{\text{cls}} k_i^T}{\sqrt{d}}\right) v_i
\end{equation}
where $q, k, v$ are the query, key, and value representations. Both Grad-CAM and Grad-CAM++ backpropagate through this same softmax, so the softmax bottleneck alone cannot explain why only Grad-CAM++ destabilizes. The difference lies in how each method converts the resulting gradients into channel weights. Grad-CAM averages the first-order gradients over all spatial locations, $\alpha_k^c = \tfrac{1}{Z}\sum_{i,j} \partial Y^c / \partial A^k_{i,j}$, and this spatial average cancels the zero-mean gradient oscillations and stays bounded. Grad-CAM++ instead weights every location by a ratio of higher-order derivatives,
\begin{equation}
\alpha_{i,j}^{kc} = \frac{\dfrac{\partial^2 Y^c}{(\partial A^k_{i,j})^2}}{2\,\dfrac{\partial^2 Y^c}{(\partial A^k_{i,j})^2} + \displaystyle\sum_{a,b} A^k_{a,b}\,\dfrac{\partial^3 Y^c}{(\partial A^k_{i,j})^3}},
\end{equation}
whose denominator mixes second and third-order terms and can pass through zero. Dividing by this small, sign-varying denominator, rather than the mere presence of higher-order terms, is what amplifies numerical noise into the channel weights.

This account predicts that the instability should track the density of the softmax coupling, and the Max-Sensitivity table (Table~\ref{tab:ms}) bears it out. The denominator is $2\,\partial^2 Y^c/(\partial A^k_{i,j})^2 + \sum_{a,b} A^k_{a,b}\,\partial^3 Y^c/(\partial A^k_{i,j})^3$. Under dense global attention, every spatial token $(a,b)$ contributes to the class logit through the shared softmax, so the third-order sum aggregates many terms that vary in sign across locations and can cancel the positive second-order term, driving the denominator toward zero. Under windowed or spatial-reduction attention, each location's higher-order derivatives depend on only its own local neighborhood, so the sum has few terms, the positive second-order term dominates, and the denominator stays bounded away from zero. Consistent with this, where attention is globally dense, the higher-order cross terms are maximized: Grad-CAM++ reaches MS $17.18$ on the isotropic ViT-B/16 against $1.94$ for Grad-CAM, and $1.74$ against $0.29$ on the globally-mixed MaxViT-S. Where attention is windowed or sparse, the coupling is local, the denominator stays away from zero, and the two methods become almost indistinguishable: on Swin-B they score $0.206$ and $0.204$, and on PVT-v2 $0.174$ and $0.182$. The collapse is therefore specific to dense global attention, not to transformers in general, and it coincides with the localization drop of Grad-CAM++ to $0.49$ on ViT-B/16 while Grad-CAM holds $0.93$, and both stay near $0.97$ on the CNN where the GAP assumption is exact.

To confirm that this divergence is a real property of the estimator and not an implementation artifact, we re-ran both CAM methods on $100$ images and recorded the full per-image Max-Sensitivity distribution (Table~\ref{tab:cam_ms}). The two methods share a single code path and differ only in the $\alpha$-weight formula, so a missing normalization step, the failure mode a reviewer would reasonably suspect, would move both equally. It does not. Grad-CAM stays tight on every backbone (median $\le 0.9$), whereas Grad-CAM++ is statistically identical to Grad-CAM under windowed (Swin-B $0.18/0.18$), spatial-reduction (PVT-v2 $0.15/0.15$), and convolutional (MobileViT $0.20/0.22$) mixing, and explodes only where attention couples all tokens at once: on ViT-B/16 its Max-Sensitivity has median $2.0$ but mean $18.1$ and a worst-image value of $523$, reproducing the $17.18$ of Table~\ref{tab:ms}, and on the global-grid MaxViT-S it fails to produce a defined map on $78$ of $100$ images. This heavy right tail, the mean sitting far above the median together with a large fraction of degenerate maps, is the gradient-variance signature the $\alpha$-denominator mechanism predicts, and it appears for Grad-CAM++ alone and only under global attention. The main tables use the \texttt{pytorch-grad-cam} backend, whereas this stability re-run is a minimal reimplementation reading from the input normalization of the last transformer block on the isotropic ViT. The absolute plain-Grad-CAM value is sensitive to that backend and target-layer choice (MS $1.94$ in Table~\ref{tab:ms} versus $0.27$ here on ViT-B/16), a reproducibility caveat in its own right, but the ordering that matters, Grad-CAM stable and Grad-CAM++ exploding under global attention, holds under both. For the heavy-tailed cells, the two tables agree once the outlier-robust median is used: EfficientViT-B1 Grad-CAM++ has median $0.71$ here against $1.17$ in Table~\ref{tab:ms}, and the $276$ mean reflects a single catastrophic image, not a separate measurement.

\begin{table*}[!t]
\centering
\caption{Grad-CAM versus Grad-CAM++ Max-Sensitivity on $100$ images, median\,(mean). The median is the robust statistic and agrees in order of magnitude with the full-grid Table~\ref{tab:ms}; the mean is shown only to expose the heavy right tail and can be dominated by a single catastrophic image (e.g., EfficientViT-B1, where one image drives the mean to $276$ while the median is $0.71$). ``GC++ max/undef.'' is the worst-image value and the count of degenerate (constant) maps. Grad-CAM++ matches Grad-CAM under local mixing and destabilizes only under dense global attention.}
\label{tab:cam_ms}
\begin{tabular}{llccc}
\toprule
Model & Attention & Grad-CAM & Grad-CAM++ & GC++ max/undef. \\
\midrule
RN-50     & CNN               & $0.26\,(0.32)$ & $0.19\,(0.21)$ & $0.5$ / $0$ \\
ViT-B/16  & global softmax    & $0.24\,(0.27)$ & $2.03\,(18.1)$ & $523$ / $12$ \\
Swin-B    & windowed          & $0.18\,(0.20)$ & $0.18\,(0.20)$ & $0.6$ / $0$ \\
PVT-v2    & spatial-reduction & $0.15\,(0.20)$ & $0.15\,(0.18)$ & $1.7$ / $0$ \\
MaxViT-S  & global grid       & $0.25\,(0.29)$ & $0.77\,(1.53)$ & $11.5$ / $78$ \\
MobileViT & conv-hybrid       & $0.20\,(0.23)$ & $0.22\,(0.25)$ & $1.7$ / $0$ \\
EffViT-B1 & linear (global)   & $0.89\,(1.49)$ & $0.71\,(276)$  & $2.3\!\times\!10^{4}$ / $16$ \\
EffViT-B2 & linear (global)   & $0.86\,(1.48)$ & $0.84\,(1.94)$ & $70$ / $10$ \\
\bottomrule
\end{tabular}

\vspace{2pt}
\raggedright\footnotesize Max-Sensitivity median\,(mean) over $100$ ImageNet-S images, $\mathrm{nr\_samples}=3$, matching the protocol of Table~\ref{tab:ms}. ViT-B/16 uses the standard ViT target layer; all other rows use the terminal feature map.
\end{table*}

\subsection{Resolution Artifacts and the Coarse CAM Grid}
\label{sec:swin_bottlenecks}
\begin{figure*}[t]
  \centering
  \includegraphics[width=0.95\textwidth]{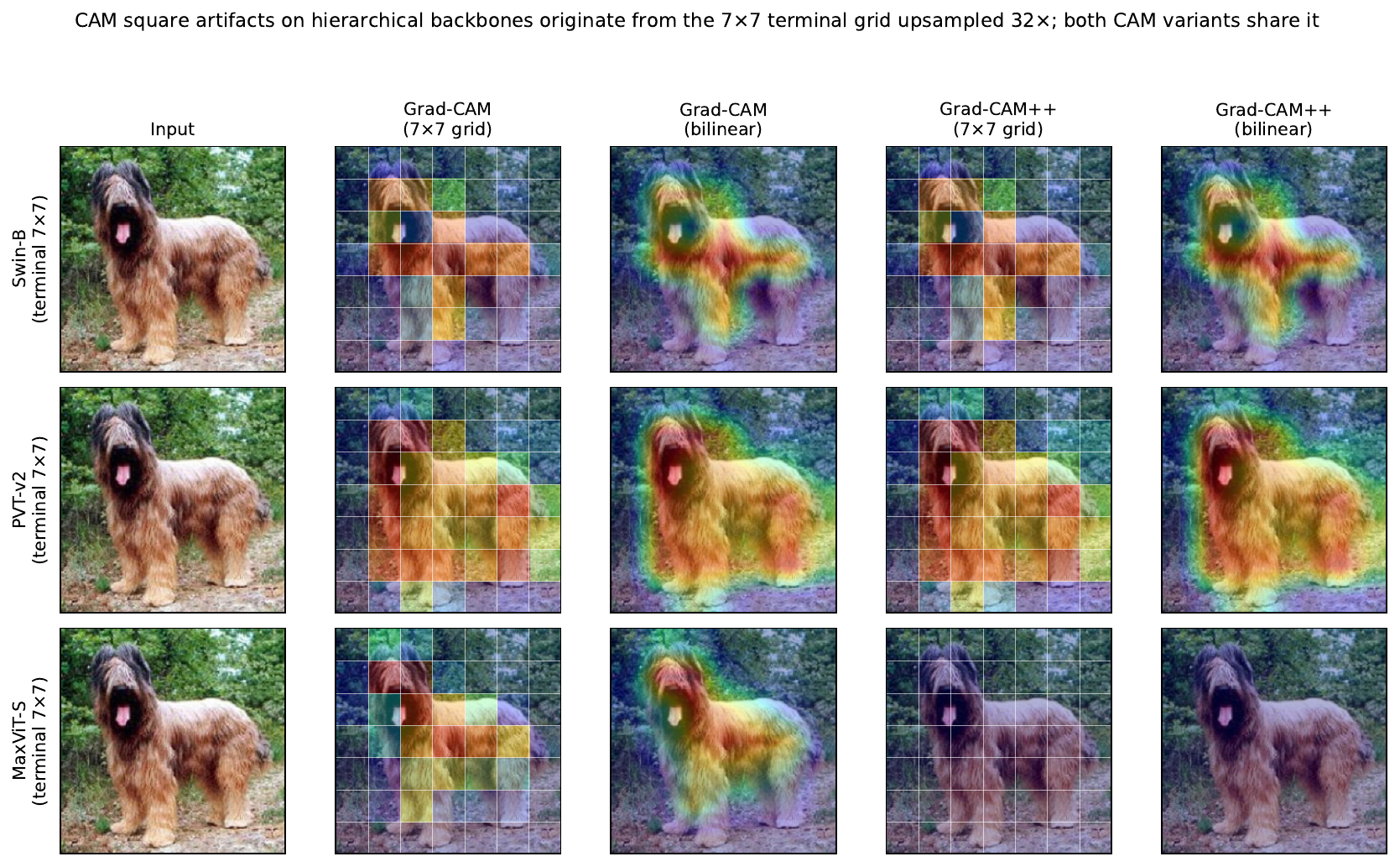}
  \caption{\textbf{The square blocks in CAM maps are a $7 \times 7$ upsampling artifact, shared by both CAM variants.} For three hierarchical backbones with a $7 \times 7$ terminal feature map, the native CAM grid (nearest-neighbor, gridlines at the cell boundaries) and the bilinear map actually used. The block edges are the $7 \times 7$ cells upsampled $32\times$; Grad-CAM and Grad-CAM++ produce the same structure because they share the terminal map. MaxViT-S Grad-CAM++ is near-empty here, an instance of the Grad-CAM++ degeneracy under global attention (Section~\ref{sec:gradcam_collapse}).}
  \label{fig:cam_grid}
\end{figure*}

The CAM maps on the hierarchical backbones carry a visible square-block structure, seen in the Swin-B and PVT-v2 Grad-CAM columns of Figure~\ref{fig:qual-grid-cat} and isolated in Figure~\ref{fig:cam_grid}. This is a resolution artifact, not a property of the object or of a particular CAM variant. Swin-B, PVT-v2, and MaxViT-S all terminate in a $7 \times 7$ feature map, so a CAM computes a $7 \times 7$ relevance grid and upsamples it $32\times$ to the $224 \times 224$ input. The block edges coincide with the $7 \times 7$ cell boundaries, and because Grad-CAM and Grad-CAM++ read the same terminal map through the same bilinear upsampling, the artifact appears identically for both (Figure~\ref{fig:cam_grid}). It is therefore not the shifted-window partitioning of Swin: PVT-v2 and MaxViT-S have no shifted windows yet show the same, or stronger, blocking.

This coarse grid bears directly on the localization results. A $7 \times 7$ blob covers the object at the granularity of $32$-pixel cells rather than tracing its silhouette, so the strong Pointing Game scores of Grad-CAM on these backbones (Table~\ref{tab:pg}) reflect a peak that lands somewhere on a large central object under a coarse blob, and should be read together with the bounding-box saturation discussed in Section~\ref{sec:quant}. Gradient-based methods, which operate at full input resolution, do not share this block structure; on Swin they instead show finer, window-aligned texture, consistent with gradient flow being channeled through local windows, though we report that as a qualitative observation rather than a measured effect.

\subsection{Limitations and Future Work}
\label{sec:limitations}

\begin{table}[t]
\centering
\caption{Evaluated model groups, spanning the phases of ViT evolution together with a pre-ViT convolutional reference (ResNet-50).}
\label{evaluated_models}
\begin{tabular}{lll}
\toprule
Model group & Representative models & Evo. Phase \\
\midrule
Standard ViT & ViT-B/16 & Genesis (2020) \\
Convolutional & ResNet-50 & Pre-ViT \\
Hierarchical & Swin, PVT-v2 variants & Phase 2 \\
Multi-axis & MaxViT variants & Phase 2 \\
Hybrid & MobileViT variants & Phase 3 \\
Efficient & EfficientViT variants & Phase 3 \\
\bottomrule
\end{tabular}
\end{table}

This study represents a rigorously controlled first step in cross-architecture XAI evaluation, and it maintains strict boundaries to isolate structural variables. In terms of scope, our controlled quantitative analysis spans eight backbones (ResNet-50, ViT-B/16, Swin-B, PVT-v2-B2, MaxViT-S, MobileViT-v2, EfficientViT-B1, and EfficientViT-B2) evaluated on a fixed set of 1,000 ImageNet-S images, covering the CNN, isotropic, hierarchical, multi-axis, hybrid, and linear-attention families of Table~\ref{evaluated_models}. Extending this grid to additional scales and further members within each family remains part of our ongoing work.

Methodologically, we excluded standard Captum LRP because its legacy formulation does not run on modern transformer backbones and rejects unsupported modules outright. We report AttnLRP only for ViT-B/16, where its published rules natively handle standard softmax self-attention; the toolbox has no rules for windowed, spatial-reduction, multi-axis, hybrid, or linear attention, so we leave those cells undefined rather than report generic-fallback numbers that would not represent AttnLRP (Section~\ref{sec:framework}). Deriving conservation-valid relevance rules for these non-standard attention mechanisms is beyond the scope of this benchmark and is a clear direction for future work.

In terms of evaluation metrics, the bounding-box Pointing Game saturates for compact, centered maps, which is why we also report the dense-mask Energy-Based Pointing Game (Table~\ref{tab:ebpg}) and treat it as the localization metric of record; extending EBPG to the perturbation-family methods, whose per-image cost is high, is the remaining step. The near-zero FC scores reflect the inherently noisy nature of standard occlusion-based proxy metrics at this scale; separating methods on faithfulness will require on-manifold perturbation baselines or a shift toward causal and localization evidence rather than single-image correlation. This FC conclusion is stated for the tested configuration (subset size $224$, $20$ runs), and a sweep over these settings would strengthen it. We release per-cell standard deviations with the raw outputs and test method separation with Bonferroni-corrected Friedman and Nemenyi procedures (Section~\ref{sec:stats}); the stochastic metrics (FC and MS) would additionally benefit from multi-seed confidence intervals, which we leave to future work. Extending the framework to causal ground-truth benchmarks such as FunnyBirds~\cite{hesse2023funnybirds}, and to self-supervised pretraining objectives such as DINOv2~\cite{oquab2024dinov2} and MAE~\cite{he2022mae} under the same fixed protocol, is a further direction. Finally, two scope limits bound the present study. Our metrics are aggregated over the $1{,}000$ images with no per-category breakdown, so failures concentrated in particular classes, such as small or texture-less objects, are folded into the aggregate, and a per-class analysis is a natural extension. And we report only automated metrics, with no human perceptual assessment of heatmap quality, so the scores are proxies for, not measurements of, human interpretability.

\section{Conclusion}
\label{sec:conclusion}
We presented a controlled benchmark that re-examines whether attribution-method rankings established on CNNs also hold on ViTs. Holding the data, the resolution, and the evaluation budget fixed across eight backbones spanning the CNN, isotropic, hierarchical, multi-axis, hybrid, and linear-attention families (ResNet-50, ViT-B/16, Swin-B, PVT-v2-B2, MaxViT-S, MobileViT-v2, and EfficientViT-B1/B2), and scoring thirteen methods on five \textsc{Quantus} metrics, we find that the rankings do not transfer cleanly. The localization ordering rearranges, robustness degrades sharply for some CAM-based and gradient-based methods on global-attention transformers, proxy FC is almost uninformative, and no method dominates across all axes. In practical terms, Grad-CAM remains a strong low-cost default for ViT localization, except on linear-attention backbones where its spatial-map assumptions break; attention rollout should be used only when stability matters more than spatial precision, and single-metric evaluations should be avoided. We release the harness, the configurations, and the raw outputs so that additional backbones, pretraining objectives, scales, and ground-truth arms can be added to this controlled snapshot.

\FloatBarrier
\bibliographystyle{elsarticle-num}
\bibliography{refs}

\end{document}